\documentclass{article}

\usepackage[preprint]{neurips_2022}

\usepackage[utf8]{inputenc} 
\usepackage[T1]{fontenc}    
\usepackage{hyperref}       
\usepackage{url,wrapfig}            
\usepackage{booktabs}       
\usepackage{amsfonts}       
\usepackage{nicefrac}       
\usepackage{microtype}      
\usepackage{xcolor}         
\usepackage[inline]{enumitem}
\usepackage{amsthm}

\usepackage{graphicx}
\usepackage{amsmath}
\usepackage{amssymb}
\usepackage{mathtools}
\usepackage{amsthm}
\pdfoutput=1

\newtheorem{prop}{Proposition}

\newtheorem{rmq}{Remark}

\DeclareMathOperator{\Diag}{diag}

\DeclareMathOperator*{\argmin}{argmin}

\newcommand{\eqdef}{:=}
\definecolor{mygreen}{rgb}{0.0, 0.7, 0.0}

\usepackage{bm}
\usepackage{algorithm2e}
\usepackage{algorithm}
\usepackage{subcaption}
\usepackage{wrapfig, framed, caption}
\usepackage{multirow}
\hypersetup{colorlinks,linkcolor={blue},citecolor={blue},urlcolor={black}}

\usepackage[capitalize,noabbrev]{cleveref}

\def\Dista{A}
\def\Distb{B}
\def\facA{A_1}
\def\facAA{A_2}
\def\facB{B_1}
\def\facBB{B_2}

\title{Unbalanced Low-rank Optimal Transport Solvers}

\author{
  Meyer Scetbon$^*$\\
  Microsoft Research\\
  \texttt{t-mscetbon@microsoft.com}\\
  \And
  Michael Klein$^{*}$\\
  Apple\\
  \texttt{michalk@apple.com}\\
  \And 
  Giovanni Palla\\
  Helmholtz Center Munich\\
  \texttt{giovanni.palla@helmholtz-muenchen.de} \\
  \And
  Marco Cuturi\\
  Apple\\
  \texttt{cuturi@apple.com}
}

\begin{document}

\maketitle

\begin{abstract}
%
The relevance of optimal transport methods to machine learning has long been hindered by two salient limitations.
First, the $O(n^3)$ computational cost of standard sample-based solvers (when used on batches of $n$ samples) is prohibitive.
Second, the mass conservation constraint makes OT solvers too rigid in practice: because they must match \textit{all} points from both measures, their output can be heavily influenced by outliers.
A flurry of recent works in OT has addressed these computational and modelling limitations, but has resulted in two separate strains of methods:
While the computational outlook was much improved by entropic regularization, more recent $O(n)$ linear-time \textit{low-rank} solvers hold the promise to scale up OT further.
On the other hand, modelling rigidities have been eased owing to unbalanced variants of OT, that rely on penalization terms to promote, rather than impose, mass conservation.
The goal of this paper is to merge these two strains, to achieve the promise of \textit{both} versatile/scalable unbalanced/low-rank OT solvers. 
We propose custom algorithms to implement these extensions for the linear OT problem and its Fused-Gromov-Wasserstein generalization, and demonstrate their practical relevance to challenging spatial transcriptomics matching problems.

\end{abstract}

\section{Introduction}
Recent machine learning (ML) works have witnessed a flurry of activity around optimal transport (OT) methods. The OT toolbox provides convenient, intuitive and versatile ways to quantify the difference between two probability measures, either to quantify a distance (the Wasserstein and Gromov-Wasserstein distances), or, in more elaborate scenarios, by computing a push-forward map that can transform one measure into the other~\citep{Peyre2019computational}. Recent examples include, e.g., single-cell omics~\citep{bunne2021learning,bunne2022supervised,SCOT2020,nitzan2019gene,cang2023screening,Klein2023.05.11.540374}, attention mechanisms~\citep{pmlr-v119-tay20a,sander2022sinkformers}, self-supervised learning\citep{caron2020unsupervised,oquab2023dinov2}, and learning on graphs~\citep{cuaz23}.

\textbf{On the challenges of using OT.} 
Despite their long presence in ML~\citep{rubner-2000}, OT methods have long suffered from various limitations, that arise from their statistical, computational, and modelling aspects. The \textit{statistical} argument is commonly referred to as the curse-of-dimensionality of OT estimators: the Wasserstein distance between two probability densities, and its associated optimal Monge map, is poorly approximated using samples as the dimension $d$ of observation grows~\citep{dudley1966weak,boissard2014mean}. On the \textit{computational} side, computing OT between a pair of $n$ samples involves solving a (generalized) matching problem, with a price of $O(n^3)$ and above~\citep{kuhn1955hungarian,ahuja1993network}. Finally, the original \textit{model} for OT rests on a mass conservation constraint: all observations from either samples must be accounted for, including outliers that are prevalent in machine learning datasets. Combined, these weaknesses have long hindered the use of OT, until a more recent generation of solvers addressed these three crucial issues.

\textbf{The Entropic Success Story.}
The winning approach, so far, to carry out that agenda has been entropic regularization methods~\citep{cuturi2013sinkhorn}. 
The computational virtues of the Sinkhorn algorithm when solving OT~\citep{altschuler2017near,peyre2016gromov,solomon2016entropic} come with statistical efficiency~\citep{genevay2018sample,mena2019statistical,chizat2020faster}, and can also be seamlessly combined with \textit{unbalanced} formulations by penalizing -- rather than constraint -- mass conversation, both for the linear \citep{FrognerNIPS,chizat2018unbalanced,sejourne2022faster,pmlr-v139-fatras21a} and quadratic~\citep{sejourne2021unbalanced} problems. These developments have all been implemented in popular OT packages~\citep{feydy2019fast,flamary2021pot,cuturi2022optimal}.

\textbf{The Low-Rank Alternative.}
A recent strain of solvers relies instead on \textit{low-rank} (LR) properties of cost and coupling matrices~\citep{forrow2018statistical,scetbon2020linear,scetbon2021lowrank}. Much like entropic solvers, these LR solvers have a better statistical outlook~\citep{scetbon2022lot} and extend to GW problems~\citep{scetbon2022linear}. In stark contrast to entropic solvers, however, LR solvers benefit from linear complexity $O(nrd)$ w.r.t sample size $n$ (using rank $r$ and cost dimension $d$) that can scale to ambitious tasks where entropic solvers fail~\citep{Klein2023.05.11.540374}. 

\textbf{The Need for Unbalanced Low-Rank Solvers.}
LR solvers do suffer, however, from a major practical limitation: their inability to handle unbalanced problems. Yet, unbalancedness is a crucial ingredient for OT to be practically relevant. This is exemplified by the fact that unbalancedness played a crucial role in the seminal reference~\citep{schiebinger2019optimal}, where it is used to model cell birth and death.

\textbf{Our Contributions}  We propose in this work to lift this last limitation for LR solvers to:
\begin{itemize}[leftmargin=.3cm,itemsep=.0cm,topsep=0cm,parsep=2pt]
	\item Incorporate unbalanced regularizers to define a LR linear solver (\S~\ref{sec:ulot1});
	\item Provide accelerated algorithms, inspired by some of the recent corrections proposed by~\citep{sejourne2022faster}, to isolate translation terms that appear in dual subroutines (\S~\ref{sec:ulot-dyk});
        \item Carry over and adapt these approaches to the GW (\S~\ref{sec:ulgw}) and Fused-GW problems (\S~\ref{sec:ulfgw});
	\item Carry out an exhaustive hyperparameter selection procedure within large scale OT tasks (spatial transcriptomics, brain imaging), and demonstrate the benefits of our approach (\S~\ref{sec:exp}).
\end{itemize}
\section{Reminders on Low-Rank Transport and Unbalanced Transport}

We consider two metric spaces $(\mathcal{X},d_{\mathcal{X}})$ and $(\mathcal{Y},d_{\mathcal{Y}})$, as well as a cost function $c:\mathcal{X}\times\mathcal{Y}\to[0,+\infty[$. The simplex $\Delta_n^{+}$ holds all positive $n$-vectors summing to $1$. For $n,m\geq 1, a\in\Delta_n^{+}$, and $b\in \Delta_m^{+}$, given points $x_1,\dots,x_n\in\mathcal{X}$ and $y_1,\dots,y_m\in\mathcal{Y}$, we define two discrete probability measures $\mu$ and $\nu$ as $\mu\eqdef\sum_{i=1}^n a_i\delta_{x_i}$, $\nu\eqdef\sum_{j=1}^m b_j\delta_{y_j}$ where $\delta_z$ is the Dirac mass at $z$.

\textbf{Cost matrices.}
For $q\geq 1$, consider first two square pairwise \textit{cost} matrices, each encoding the geometries of points \textit{within} $\mu$ and $\nu$, and a rectangular matrix that studies that \textit{across} their support:
$$\Dista:=[d_{\mathcal{X}}^q(x_i,x_{i'})]_{1 \leq i,i'\leq n},\; \Distb:=[d_{\mathcal{Y}}^q(y_j,y_{j'})]_{1 \leq j,j'\leq m}\,,\; C\eqdef [c(x_i,y_j)]_{1\leq i,j\leq n,m}\,.$$ 

\textbf{The Kantorovich Formulation of OT} is defined as the following linear program, defined by $C$:
\begin{align}
\label{eq-ot}
    \text{OT}(\mu,\nu) \eqdef \min_{P\in\Pi_{a,b}}\langle C,P\rangle\,,\quad \text{where}\quad \Pi_{a,b}\eqdef\left\{P\in\mathbb{R}_{+}^{n\times m},~\text{s.t.}~~P\mathbf{1}_m=a,~P^T\mathbf{1}_n=b\right\}\,.
\end{align}

\textbf{The Low-Rank Formulation of OT} is best understood as a variant of \eqref{eq-ot} that rests on a low-rank \textit{property} for cost matrix $C$, and low-rank \textit{constraints} for couplings $P$. More precisely, \citet{scetbon2021lowrank} propose to constraint the set of admissible couplings to those, within $\Pi_{a,b}$, that have a non-negative rank of $r\geq 1$.
That set can be equivalently reparamaterized as 
\begin{align*}
\Pi_{a,b}(r) =\{P\in\mathbb{R}^{n\times m}_+ | P = Q \Diag(1/g)R^T,~~Q\in\Pi_{a,g},~~ R\in\Pi_{b,g},~~\text{and}~~ g\in\Delta_r^+\}.
\end{align*}
The low-rank optimal transport (LOT) problem simply uses that restriction in~\eqref{eq-ot} to define :
\begin{align}
\label{eq-lot}
\text{LOT}_r(\mu,\nu) \eqdef \min_{P\in\Pi_{a,b}(r)}\langle C,P \rangle = \min_{Q\in\Pi_{a,g}, R\in\Pi_{a,g}, g\in\Delta^+_r}  \langle C,Q\Diag(g)R\rangle\,.\end{align}
\citet{scetbon2021lowrank} propose and prove the convergence of a mirror-descent scheme to solve~\eqref{eq-lot}, and obtain linear time and memory complexities with respect to the number of samples, where each iteration in that descent scales as $(n+m)rd$, where $d$ is the rank of $C$.

\textbf{The Unbalanced Formulation of OT} starts from~\eqref{eq-ot} as well, but proposes to do without $\Pi_{a,b}$ and its marginal constraints~\citep{FrognerNIPS,chizat2018unbalanced}, and rely instead on two regularizers:
\begin{equation}
\label{eq-uot}
    \text{UOT}(\mu,\nu) \eqdef  \min_{P\in\mathbb{R}^{n\times m}_+}\langle C,P\rangle + \tau_1 \textrm{KL}(P\mathbf{1}_m|a) + \tau_2 \textrm{KL}(P^T\mathbf{1}_n|b).
\end{equation}
This formulation is solved using entropic regularization, with modified Sinkhorn updates ~\citep{FrognerNIPS}. \textit{Proposing an efficient algorithm able to merge \eqref{eq-lot} with \eqref{eq-uot} is the first goal of this paper.}

\textbf{Gromov-Wasserstein (GW) Considerations.} The GW problem~\citep{memoli-2011} is a generalization of~\eqref{eq-ot} where the energy $\mathcal{Q}_{\Dista, \Distb}$ is a quadratic function of $P$ defined through inner cost matrices $A$, $B$:
\begin{align}
\label{eq-obj-GW}
\!\!\mathcal{Q}_{\Dista, \Distb}(P)\!\eqdef \!\!\!\!\!\sum_{i,j,i',j'}\!\! (\Dista_{ii'} - \Distb_{jj'})^2 P_{ij}P_{i'j'}\! =\! \mathbf{1}_m^TP^T \Dista^{\odot2} P\mathbf{1}_m +  \mathbf{1}_n^T P \Distb^{\odot2}P^T \mathbf{1}_n -2\langle \Dista P\Distb,P\rangle\; .
\end{align}
To minimize \eqref{eq-obj-GW}, the default approach rests on entropic regularization~\citep{solomon2016entropic,peyre2016gromov} and variants~\citep{sato2020fast,blumberg2020mrec,NEURIPS2019_6e62a992,liconvergent}. \citet{scetbon2022linear} adapted the low-rank framework to minimize $\mathcal{Q}_{\Dista, \Distb}$ over low-rank matrices $P$, achieving a linear-time complexity when $A$ and $B$ are themselves low-rank. Independently, \citep{sejourne2021unbalanced} proposed an unbalanced generalization that also applies to GW and which can be implemented practically using entropic regularization. Finally, the minimization of a composite objective involving both $\mathcal{Q}_{\Dista, \Distb}$ and $\langle C,\cdot \rangle$ is known as the \textit{fused} GW problem~\citep{vayer2018fused}.
\section{Unbalanced Low-Rank Transport}\label{sec:ulot}
\subsection{Unbalanced Low-rank Linear Optimal Transport}\label{sec:ulot1}

We incorporate unbalancedness to low-rank solvers~\citep{scetbon2021lowrank,scetbon2022linear}, moving gradually from the linear problem to the more involved GW and FGW problem. Using the framework of~\citep{FrognerNIPS,chizat2018unbalanced}, we can first extend the definition of LOT, introduced in~\eqref{eq-lot}, to the unbalanced case by considering the following optimization problem:
\begin{equation}
\label{eq-ulot}
\begin{aligned}
    \text{ULOT}_r(\mu,\nu) \eqdef  \min_{P\text{:}~\text{rk}_+(P)\leq r}\langle C,P\rangle + 
    &\tau_1 \text{KL}(P\mathbf{1}_m|a) + \tau_2 \text{KL}(P^T\mathbf{1}_n|b),
\end{aligned}
\end{equation}
where $\text{rk}_+(P)$ denotes the nonnegative rank of $P$. Therefore by denoting $\Pi_r\eqdef \{(Q,R,g)\in\mathbb{R}_{+}^{n\times r}\times\mathbb{R}_{+}^{m\times r}\times\mathbb{R}_{+}^{r}\text{:}~ Q^T \mathbf{1}_n =R^T \mathbf{1}_m = g \}$, and using the repamatrization of low-rank couplings, we obtain the following equivalent formulation of ULOT:

\begin{equation}
\label{eq-ulot-reformulated}
\begin{aligned}
  \text{ULOT}_r(\mu,\nu) = \min_{(Q,R,g)\in \Pi_r} \underbrace{\langle C,Q \Diag(1/g)R^T\rangle}_{\mathcal{L}_C(Q,R,g)}
   +\underbrace{\tau_1 \text{KL}(Q\mathbf{1}_r|a) + \tau_2 \text{KL}(R\mathbf{1}_r|b)}_{\mathcal{G}_{a,b}(Q,R,g)}\; .
\end{aligned}
\end{equation}

We introduce slightly more compact notations for $\mathcal{G}_{a,b}(Q,R,g) = F_{\tau_1,a}(Q\mathbf{1}_r)+F_{\tau_2,b}(R\mathbf{1}_r),$
where $F_{\tau,z}(s)=\tau \text{KL}(s|z)$ for $\tau>0$ and $z\geq 0$ coordinate-wise. To solve~\eqref{eq-ulot-reformulated}, and using this split, we move away from mirror-descent and apply instead proximal gradient-descent with respect to the KL divergence. At each iteration, we consider a linear approximation of $\mathcal{L}_C$ where a $\text{KL}$ penalization is added to the objective, as in the classical mirror descent, however, we leave $\mathcal{G}_{a,b}$ intact at each iteration. Borrowing notations from~\citep{scetbon2021lowrank}, we must solve at each iteration the convex optimization problem:
\begin{equation}
\label{eq-barycenter-ulot} 
\begin{aligned}
 (Q_{k+1},R_{k+1},g_{k+1})   \eqdef   \argmin_{\bm{\zeta} \in\Pi_r} \frac{1}{\gamma_k}\text{KL}(\bm{\zeta},\bm{\xi}_k)+
 \tau_1 \text{KL}(Q\mathbf{1}_r|a) + \tau_2 \text{KL}(R\mathbf{1}_r|b)
 \end{aligned}
\end{equation}
where $(Q_0,R_0,g_0)\in\Pi_r$ is an initial point,
$\bm{\xi}_k \eqdef (\xi_{k}^{(1)},\xi_{k}^{(2)},\xi_{k}^{(3)})$ holds running costs matrices defined as 
$$\xi_{k}^{(1)} \eqdef Q_k \odot e^{-\gamma_kCR_k \Diag(1/g_k)}, \xi_{k}^{(2)} \eqdef R_k \odot e^{-\gamma_kC^TQ_k \Diag(1/g_k))}, \xi_{k}^{(3)} \eqdef g_k \odot e^{\gamma_k\omega_k/g_k^2},$$
 with $[\omega_k]_i \eqdef [Q_k^TCR_k]_{i,i}$ for all $i\in\{1,\dots,r\}$, and $(\gamma_k)_{k\geq 0}$ is a sequence of positive step sizes.
 
\textbf{Reformulation using Duality.} To solve~\eqref{eq-barycenter-ulot}, we follow~\citep{scetbon2021lowrank} and apply Dykstra’s algorithm~\citep{dykstra1983algorithm}. The iterations of this algorithm takes a very simple form that can be obtained as an alternating maximization on the dual formulation of~\eqref{eq-barycenter-ulot}, provided as follows.
\begin{prop}
\label{prop-dual}
    The convex optimization problem defined in~\eqref{eq-barycenter-ulot} admits the following dual:
    \begin{equation}
      \label{eq-dual-barycenter}
    \begin{aligned}
        \sup_{f_1,h_1,f_2,h_2} & \mathcal{D}_k(f_1,h_1,f_2,h_2) := - F_{\tau_1,a}^{*}(-f_1) - \frac{1}{\gamma_k}\langle e^{\gamma_k(f_1\oplus h_1)} -1, \xi^{(1)}_k\rangle \\
        &- F_{\tau_2,b}^{*}(-f_2) - \frac{1}{\gamma_k} \langle e^{\gamma_k(f_2\oplus h_2)} -1, \xi^{(2)}_k\rangle - \frac{1}{\gamma_k}\langle e^{- \gamma_k(h_1 + h_2)} - 1,\xi^{(3)}_k\rangle 
    \end{aligned}
    \end{equation}
    where $h_1,h_2\in\mathbb{R}^r$, $f_1\in\mathbb{R}^n$, $f_2\in\mathbb{R}^m$, $ F_{\tau,z}^{*}(\cdot)\eqdef \sup_{y}\{\langle y, \cdot\rangle - F_{\tau,z}(y)\}$ is the convex conjugate of $ F_{\tau,z}$. In addition strong duality holds and the primal problem admits a unique minimizer.
\end{prop}
\begin{rmq}
    While we stick to KL regularizers in this work for simplicity, it is worth noting that this can be extended to more generic regularizers $F_{\tau_1,a}$ and  $F_{\tau_2,b}$, as considered by~\citet{chizat2018unbalanced}.
\end{rmq}
We use an alternating maximization scheme to solve ~\eqref{eq-dual-barycenter}. Starting from $h_1^{(0)}=h_2^{(0)}=\mathbf{0}_r$, we apply for $\ell\geq 0$ the following updates (dropping iteration number $k$ in~\eqref{eq-barycenter-ulot} for simplicity):
$$
\begin{aligned}
     f_1^{(\ell+1)}\eqdef\arg\sup_{z} \mathcal{D}(z,h_1^{(\ell)},f_2^{(\ell)},h_2^{(\ell)}), \,f_2^{(\ell+1)}\eqdef\arg\sup_{z} \mathcal{D}(f_1^{(\ell+1)},h_1^{(\ell)},z,h_2^{(\ell)}), \\
     (h_1^{(\ell+1)},h_2^{(\ell+1)})\eqdef \arg\sup_{z_1,z_2} \mathcal{D}(f_1^{(\ell+1)},z_1,f_2^{(\ell+1)},z_2).
\end{aligned}
$$
These maximizations can all be obtained in closed form, to result in the closed-form updates:
\begin{align*}
     &\exp(\gamma f_1^{(\ell+1)}) = \left(\frac{a}{\xi^{(1)}\exp(\gamma h_1^{(\ell)})}\right)^{\frac{\tau_1}{\tau_1+1/\gamma}},\quad
      \exp(\gamma f_2^{(\ell+1)}) = \left(\frac{b}{\xi^{(2)}\exp(\gamma h_2^{(\ell)})}\right)^{\frac{\tau_2}{\tau_2+1/\gamma}}\\
    &g_{\ell+1} \eqdef \left(\xi^{(3)}\odot  (\xi^{(1)})^T \exp(\gamma f_1^{(\ell+1)}) \odot (\xi^{(2)})^T \exp(\gamma f_2^{(\ell+1)})\right)^{1/3}\\
    &\exp(\gamma h_1^{(\ell+1)}) = \frac{g_{\ell+1}}{(\xi^{(1)})^T \exp(\gamma f_1^{(\ell+1)})},\quad \exp(\gamma h_2^{(\ell+1)}) = \frac{g_{\ell+1}}{(\xi^{(2)})^T \exp(\gamma f_2^{(\ell+1)})}
\end{align*}
When using ``scaling'' representations for these dual variables, $\ell\geq 0$, $u^{(\ell)}_{i}\eqdef \exp(\gamma f_i^{(\ell)})$ and  $v^{(\ell)}_{i}\eqdef \exp(\gamma h_i^{(\ell)})$ for $i\in\{1,2\}$, we obtain a simple update, provided in the appendix (Alg.~\ref{alg-dykstra-ulot}).

\textbf{Initialization and Termination.} We use the stopping criterion proposed in~\citep{scetbon2021lowrank} to terminate the algorithm,
$\Delta(\bm{\zeta},\bm{\tilde{\zeta}},\gamma) \eqdef \frac{1}{\gamma^2}(\mathrm{KL}(\bm{\zeta},\bm{\tilde{\zeta}})+\mathrm{KL}(\bm{\tilde{\zeta}},\bm{\zeta}))$. Combined with practical improvements proposed in~\citep{scetbon2022lot} to initialize the algorithm, and adapt the choice of $\gamma_k$ at each iteration $k$ of the outer loop,
we can now summarize our proposal in Algorithm~\ref{alg-prox-ulot}, which can be seen as an extension of~\citep[Alg.2]{scetbon2021lowrank}.

\textbf{Convergence and Complexity.} The proof of convergence of the Dykstra algorithm (Alg.~\ref{alg-dykstra-ulot}) can be found for example in~\citep{BauschkeCombettes-Dykstra}). In addition,~\citep{scetbon2021lowrank} show the convergence of their scheme towards a stationary points w.r.t to the criterion $\Delta(\cdot,\cdot,\gamma)$ for $\gamma$ fixed along the iterations of the outer loop. In terms of complexity, given $\bm{\xi}$, solving Eq.~\eqref{eq-barycenter-ulot} requires a time and memory complexity of $\mathcal{O}((n+m)r)$. However computing  $\bm{\xi}$ requires in general $\mathcal{O}((n^2+m^2)r)$ time and  $\mathcal{O}(n^2+m^2)$ memory. In \citep{scetbon2021lowrank}, the authors propose to consider low-rank approximation of the cost matrix $C$ of the form $C\simeq C_1C_2^T$  where $C_1\in\mathbb{R}^{n\times d}$ and $C_2\in\mathbb{R}^{m\times d}$: in that case computing $\bm{\xi}$ can be done in $\mathcal{O}((n+m)rd)$ time and $\mathcal{O}((n+m)(r+d))$ memory. Such approximations can be obtained using the algorithm in~\citep{indyk2019sampleoptimal} which guarantees that for any distance matrix $C\in\mathbb{R}^{n\times m}$ and $\alpha>0$ it can outputs matrices $C_1\in\mathbb{R}^{n\times d}$, $C_2\in\mathbb{R}^{m\times d}$ in $\mathcal{O}((m+n)\text{poly}(\frac{d}{\alpha}))$ algebraic operations such that with probability at least $0.99$ that $\Vert C - C_1C_2^T\Vert_F^2\leq \Vert C - C_d\Vert_F^2 +\alpha\Vert C\Vert_F^2$, where $C_d$ denotes the best rank-$d$ approximation to $C$.

\begin{algorithm}[H]
\SetAlgoLined
\textbf{Inputs:} $C, a, b, r, \gamma_0, \tau_1,\tau_2,\delta$\\
$Q,R,g \gets \text{Initialization as proposed in~\citep{scetbon2022lot}}$\\
\Repeat{$\Delta((Q,R,g),(\tilde Q,\tilde R,\tilde g),\gamma)<\delta$}{
    $\tilde{Q}=Q,~~\tilde{R}=R,~~\tilde{g}=g,\\
    \nabla_Q = CR \Diag(1/g),~~\nabla_R = C^\top Q \Diag(1/g),\\
    \omega \gets \mathcal{D}(Q^TCR),~~\nabla_g = - \omega / g^2,\\
    \gamma \gets \gamma_0 / \max(\|\nabla_Q\|_{\infty}^2,\|\nabla_R\|_{\infty}^2, \|\nabla_g\|_{\infty}^2),\\
    \xi^{(1)}\gets Q \odot \exp(-\gamma\nabla_Q),~\xi^{(2)}\gets R\odot \exp(-\gamma\nabla_R),~\xi^{(3)}\gets g\odot \exp(-\gamma\nabla_g),\\
    Q,R,g\gets \text{ULR-Dykstra}(a,b,\bm{\xi},\gamma,\tau_1,\tau_2,\delta)~(\text{Alg.~\ref{alg-dykstra-ulot})}$
  }
\textbf{Result:} $Q,R,g$
\caption{$\text{ULOT}(C,a,b,r,\gamma_0,\tau_1,\tau_2,\delta)$ \label{alg-prox-ulot}}
\end{algorithm}

\subsection{Improvements on the Unbalanced Dykstra Algorithm}\label{sec:ulot-dyk}
A well documented source of instability of unbalanced formulations of OT lies in capturing efficiently what optimal mass is targeted by such formulations.~\citet{sejourne2022faster} have proposed a technique to address this issue and lower significantly computational costs. They propose first a dual objective that is \emph{translation} invariant. 
We take inspiration from this strategy and adapt to our problem, to propose the following variant of~\eqref{eq-dual-barycenter}:
\begin{equation}
\label{eq-dual-reformulated}
\sup_{\tilde f_1, \tilde h_1, \tilde f_2, \tilde h_2} \left(\mathcal{D}_{\textrm{TI}}(\tilde f_1, \tilde h_1, \tilde f_2, \tilde h_2)\eqdef \sup_{\lambda_1,\lambda_2\in\mathbb{R}} \mathcal{D}(\tilde f_1+\lambda_1,\tilde h_1-\lambda_1,\tilde f_2+\lambda_2,\tilde h_2-\lambda_2)\right)
\end{equation}
It is clear from the reparametrization that both problems \eqref{eq-dual-barycenter} and \eqref{eq-dual-reformulated} have the same value and also that $(\tilde f_1,\tilde h_1,\tilde f_2,\tilde h_2)$ is solution of~\eqref{eq-dual-reformulated} if and only if $(\tilde f_1 +\lambda_1^*,\tilde h_1-\lambda_1^*, \tilde f_2+\lambda_2^*, \tilde h_2-\lambda_2^*)$ is solution of~\eqref{eq-dual-barycenter} where $(\lambda_1^*,\lambda_2^*)$ solves $\mathcal{D}_{\textrm{TI}}(\tilde f_1,\tilde h_1,\tilde f_2,\tilde h_2)$. To solve~\eqref{eq-dual-reformulated}, we show that the variational formulation of the translation invariant dual objective targeted inside~\eqref{eq-dual-reformulated} can be obtained in closed form.
\begin{prop}
\label{prop-TI}
    Let $  \tilde f_1\in\mathbb{R}^n$, $ \tilde f_2\in\mathbb{R}^m$ and $\tilde h_1,\tilde h_2\in\mathbb{R}^{r}$, then the inner problem defined in~\eqref{eq-dual-reformulated} by $\mathcal{D}_{\textrm{TI}}(\tilde f_1, \tilde h_1, \tilde f_2, \tilde h_2)$ admits a unique solution $(\lambda_1^\star,\lambda_2^\star)$ and we have that
    \begin{align}
        \lambda_1^\star&\eqdef \left(1- \frac{\tau_1\tau_2}{(1/\gamma +\tau_1)(1/\gamma+\tau_2)}\right)^{-1}\left(\frac{\tau_1/\gamma}{1/\gamma+\tau_1}c_1 - \frac{\tau_1/\gamma}{1/\gamma+\tau_1}\frac{\tau_2}{1/\gamma+\tau_2} c_2\right)\label{eq:lambda1}\\
        \lambda_2^\star&\eqdef \left(1- \frac{\tau_1\tau_2}{(1/\gamma +\tau_1)(1/\gamma+\tau_2)}\right)^{-1}\left(\frac{\tau_2/\gamma}{1/\gamma+\tau_2}c_2 - \frac{\tau_1/\gamma}{1/\gamma +\tau_1}\frac{\tau_2}{1/\gamma+\tau_2} c_1\right)\label{eq:lambda2}
    \end{align}
where
\begin{align*}
    c_1\eqdef \log\left( \frac{\langle\exp(-  \tilde f_1/\tau_1),a\rangle}{\langle \exp(- \gamma ( \tilde h_1+ \tilde h_2)),\xi^{(3)}\rangle}\right),\quad\text{and}\quad
       c_2\eqdef \log\left( \frac{\langle\exp(-  \tilde f_2/\tau_2),a\rangle}{\langle \exp(- \gamma ( \tilde h_1+ \tilde h_2)),\xi^{(3)}\rangle}\right).
\end{align*}
\end{prop}

We are now ready to perform an alternate maximization scheme on the translation invariant formulation of the dual $\mathcal{D}_{\textrm{TI}}$. Indeed using Danskin's theorem (under the assumption that $\lambda_1^*,\lambda_2^*$ do not diverge), one obtains a variant of Algorithm~\ref{alg-dykstra-ulot}, summarized in Algorithm~\ref{alg-dykstra-ti-ulot}.

\begin{algorithm}[H]
\SetAlgoLined
\textbf{Inputs:} 
$a,b,\xi^{(3)},u_1,v_1,u_2,v_2,\gamma,\tau_1,\tau_2\\
\tilde u_1 \gets u_1^{-1/\gamma/\tau_1},~\tilde u_2 \gets u_2^{-1/\gamma/\tau_2}\\
c_1\gets \log(\langle \tilde u_1,a \rangle) - \log(\langle\xi^{(3)},v_1^{-1}\odot v_2^{-1}\rangle),~~
c_2\gets \log(\langle \tilde u_2,b \rangle) - \log(\langle\xi^{(3)},v_1^{-1}\odot v_2^{-1}\rangle)$\\
\textbf{Result:} $\lambda_1^\star,~~\lambda_2^\star$ as in \eqref{eq:lambda1}, \eqref{eq:lambda2}
\caption{$\text{compute-lambdas}(a,b,\xi^{(3)},u_1,v_1,u_2,v_2,\gamma,\tau_1,\tau_2)$~\label{alg-compute-lamb}}
\end{algorithm}

\begin{algorithm}[H]
\SetAlgoLined
\textbf{Inputs:} $a, b,\bm{\xi}=(\xi^{(1)}, \xi^{(2)},\xi^{(3)}),\gamma,\tau_1,\tau_2,\delta$\\
$v_1=v_2=\mathbf{1}_r$, $u_1=\mathbf{1}_n$, $u_2=\mathbf{1}_m$\\
\Repeat{$\frac{1}{\gamma}\max(\Vert \log(u_i/\tilde{u}_i)\|_{\infty}, \|\log(v_i/\tilde{v}_i)\|_{\infty})<\delta$}{
    $\tilde{v}_1=v_1,~\tilde{v}_2=v_2, \tilde{u}_1 = u_1, \tilde{u}_2 = u_2\\
    \lambda_1, \lambda_2 \gets \text{compute-lambdas}(a,b,\xi^{(3)},u_1,v_1,u_2,v_2,\gamma,\tau_1,\tau_2)~(\text{Alg.~\ref{alg-compute-lamb})}\\
    u_{1} = \left(\frac{a}{\xi^{(1)}v_{1}}\right)^{\frac{\tau_1}{\tau_1+1/\gamma}}\exp(-\lambda_1/\tau_1)^{\frac{\tau_1}{1/\gamma +\tau_1}}, \quad u_{2} = \left(\frac{b}{\xi^{(2)}v_{2}}\right)^{\frac{\tau_2}{\tau_2+1/\gamma}}\exp(-\lambda_2/\tau_2)^{\frac{\tau_2}{1/\gamma +\tau_2}},\\
    \lambda_1, \lambda_2 \gets \text{compute-lambdas}(a,b,\xi^{(3)},u_1,v_1,u_2,v_2,\gamma,\tau_1,\tau_2)~(\text{Alg.~\ref{alg-compute-lamb})}\\
   g = \exp(\gamma(\lambda_1+ \lambda_2))^{1/3} \left(\xi^{(3)}\odot  (\xi^{(1)})^T u_1 \odot (\xi^{(2)})^T u_2\right)^{1/3},~v_1 = \frac{g}{(\xi^{(1)})^T u_1},~ v_2 = \frac{g}{(\xi^{(2)})^T u_2}$
  }
\textbf{Result:} $\Diag(u_1)\xi^{(1)}_k\Diag(v_1),~~\Diag(u_2)\xi^{(2)}_k\Diag(v_2),~~g$
\caption{$\text{ULR-TI-Dykstra}(a,b,\bm{\xi},\gamma,\tau_1,\tau_2,\delta)$~\label{alg-dykstra-ti-ulot}}
\end{algorithm}

\subsection{Unbalanced Low-rank Gromov-Wasserstein}\label{sec:ulgw}
The low-rank Gromov-Wasssertein (LGW) problem~\citep{scetbon2022linear} between the two discrete metric measure spaces $(\mu, d_\mathcal{X})$ and $(\nu, d_\mathcal{Y})$, written for compactness using $(a,\Dista)$ and $(b,\Distb)$, reads
\begin{align}
\label{eq-GW}
    &\text{LGW}_r((a, \Dista),(b, \Distb))=\!\!\min_{P\in\Pi_{a,b}(r)}\!\!\mathcal{Q}_{\Dista, \Distb}(P),
\end{align}
Following \S~\ref{sec:ulot1}, we introduce the unbalanced low-rank Gromov-Wasserstein problem (ULGW). There is, however, an important difference with~\eqref{eq-GW}: When $P$ is constrained to be in $\Pi_{a,b}$, the first two terms of the RHS in~\eqref{eq-GW} simplify to $a^TA^{\odot 2}a + b^TB^{\odot 2}b$. Hence, they are constant and can be discarded when optimizing. In an unbalanced setting, these terms vary and must be accounted for.
\begin{equation}
\label{eq-unbalanced-lgw}
\begin{aligned}
&\text{ULGW}_r((a, \Dista),(b, \Distb))=\!\!\min_{(Q,R,g)\in\Pi_r}\!\!
\langle \Dista^{\odot2} Q\mathbf{1}_r,Q\mathbf{1}_r\rangle +  \langle \Distb^{\odot2}R \mathbf{1}_r,R \mathbf{1}_r\rangle\\
&-2\langle \Dista Q\Diag(1/g)R^T \Distb, Q\Diag(1/g)R^T\rangle+ \tau_1\text{KL}(Q\mathbf{1}_r|a) + \tau_2\text{KL}(R\mathbf{1}_r|b)
\end{aligned}
\end{equation}
To solve the problem, we apply the same scheme as proposed for ULOT, that is a proximal gradient descent where we linearize $\mathcal{Q}_{\Dista, \Distb}$ and add a KL penalization while leaving the soft marginal constraints unchanged. Therefore the algorithm to solve ULGW is the exact same as the one solving ULOT, however, the kernels $\bm{\xi}_k$ now take into account the quadratic scores of the original LGW problem. More formally at each iteration $k$ of the outer loop, we propose to solve
\begin{equation}
\label{eq-barycenter-ulgw} 
\begin{aligned}
 (Q_{k+1},R_{k+1},g_{k+1})   \eqdef   \argmin_{\bm{\zeta} \in\Pi_r} \frac{1}{\gamma_k}\text{KL}(\bm{\zeta},\bm{\xi}_k)+
 \tau_1 \text{KL}(Q\mathbf{1}_r|a) + \tau_2 \text{KL}(R\mathbf{1}_r|b)
 \end{aligned}
\end{equation}
where $(Q_0,R_0,g_0)\in\Pi_r$ is an initial point, $(\gamma_k)_{k\geq 0}$ is a sequence of positive step sizes, $P_k=Q_k\Diag(1/g_k)R_k^T$,  $\bm{\xi}_k \eqdef (\xi_{k}^{(1)},\xi_{k}^{(2)},\xi_{k}^{(3)})$ and
\begin{align*}
    \xi_{k}^{(1)} &\eqdef Q_k \odot \exp(-2\gamma_k Q_k\mathbf{1}_r\mathbf{1}_r^T)  \odot \exp(-4\gamma_k \Dista P_k\Distb R_k\Diag(1/g_k)))\\
      \xi_{k}^{(2)} &\eqdef R_k \odot \exp(-2\gamma_k R_k\mathbf{1}_r\mathbf{1}_r^T)  \odot \exp(-4\gamma_k \Distb P_k^T\Dista Q_k\Diag(1/g_k)))\\
      \xi_{k}^{(3)}&\eqdef g_k
      \odot\exp(4\gamma_k\omega_k/g_k^2)\quad \text{with}\quad [\omega_k]_i \eqdef [Q_k^T\Dista P_k\Distb R_k]_{i,i} ~~\forall~ i\in\{1,\dots,r\}.
\end{align*}
Note that~\eqref{eq-barycenter-ulgw} is the exact same optimization problem as ~\eqref{eq-barycenter-ulot}, where only $\bm{\xi}_k$ has changed and therefore can be solved using Algorithm~\ref{alg-dykstra-ti-ulot}. Algorithm~\ref{alg-prox-ulgw} summarizes our strategy to solve~\eqref{eq-unbalanced-lgw}.

\begin{algorithm}[H]
\SetAlgoLined
\textbf{Inputs:} $\Dista, \Distb, a, b, r, \gamma_0, \tau_1,\tau_2,\delta$\\
$Q,R,g \gets \text{Initialization as proposed in~\citep{scetbon2022lot}}$\\
\Repeat{$\Delta((Q,R,g),(\tilde Q,\tilde R,\tilde g),\gamma)<\delta$}{
    $\tilde{Q}=Q,~~\tilde{R}=R,~~\tilde{g}=g,\\
    \nabla_Q = 4\Dista Q\Diag(1/g)R^T\Distb R\Diag(1/g) + 2Q\mathbf{1}_r\mathbf{1}_r^T,\\
    \nabla_R = 4\Distb R\Diag(1/g)Q^T \Dista Q\Diag(1/g) + 2R\mathbf{1}_r\mathbf{1}_r^T,\\
    \omega \gets \mathcal{D}(Q^T\Dista Q\Diag(1/g)R^T \Distb R),~~\nabla_g = - \omega / g^2,\\
    \gamma \gets \gamma_0 / \max(\|\nabla_Q\|_{\infty}^2,\|\nabla_R\|_{\infty}^2, \|\nabla_g\|_{\infty}^2),\\
    \xi^{(1)}\gets Q \odot \exp(-\gamma\nabla_Q),~\xi^{(2)}\gets R\odot \exp(-\gamma\nabla_R),~\xi^{(3)}\gets g\odot \exp(-\gamma_k\nabla_g),\\
    Q,R,g\gets \text{ULR-TI-Dykstra}(a,b,\bm{\xi},\gamma,\tau_1,\tau_2,\delta)~(\text{Alg.~\ref{alg-dykstra-ti-ulot})}$
  }
\textbf{Result:} $Q,R,g$
\caption{$\text{ULGW}(\Dista,\Distb,a,b,r,\gamma_0,\tau_1,\tau_2,\delta)$ \label{alg-prox-ulgw}}
\end{algorithm}

\textbf{Convergence and Complexity.} Similarly to linear ULOT, the unbalanced Dykstra algorithm is guaranteed to converge~\citep{bauschke2000dykstras}. in addition,~\citep{scetbon2022linear} prove the convergence of their scheme to a stationary point of the problem. Concerning the complexity, as we use Algorithm~\ref{alg-dykstra-ulot} with the same complexity, we obtain therefore the exact same complexity in terms of time of memory to solve these inner problems. The slight variation in kernel $\bm{\xi}$ compared to ULOT still retains the same $\mathcal{O}((n^2+m^2)r)$ time and $\mathcal{O}(n^2+m^2)$ memory complexities. However as in ULOT, we can take advantage of low-rank approximations of the costs matrices $\Dista$ and $\Distb$ in order to reach linear complexity. Indeed, assuming $\Dista\simeq \facA \facAA^T$ and $\Distb\simeq \facB \facBB$   where $\facA,\facAA\in\mathbb{R}^{n\times d_X}$ and  $\facB,\facBB\in\mathbb{R}^{m\times d_Y}$, then the total time and memory complexities become respectively $\mathcal{O}(mr(r+d_Y) + nr(r + d_X))$ and $\mathcal{O}((n+m)(r+d_X+d_Y))$. Again, when $\Dista$ and $\Distb$ are distance matrices, we use the algorithms from~\citep{indyk2019sampleoptimal}.

\subsection{Unbalanced Low-rank Fused-Gromov-Wasserstein}\label{sec:ulfgw}
We finally focus on the increasingly popular~\citep{Klein2023.05.11.540374} Fused-Gromov-Wasserstein problem, which merges linear and quadratic objectives~\citep{vayer2018fused}:
\begin{equation}
\label{eq-fgw}
\begin{aligned}
    \text{FGW}(\mu,\nu) &\eqdef \min_{P\in \Pi_{a,b}}\alpha\langle C , P\rangle + \bar\alpha\mathcal{Q}_{\Dista,\Distb}(P)
\end{aligned}
\end{equation}
where $\alpha\in[0,1]$ and $\bar\alpha\eqdef1-\alpha$ allows interpolating between the GW and linear OT geometries. This problem remains a GW problem, where one replaces the 4-way cost $M[i,i',j,j']\eqdef(A_{i,i'}-B_{j,j'})^2$ appearing in~\eqref{eq-obj-GW} by a composite interpolated cost between the OT and GW geometries, redefined as $M[i,i',j,j'] = \alpha C_{i,j} + \bar\alpha(A_{i,i'}-B_{j,j'})^2$. Our proposed unbalanced and low-rank version of the FGW problem includes $|P|:=\|P\|_1$ the mass of $P$, to homogenize linear and quadratic terms,
\begin{equation}
\label{eq-ufgw}
\begin{aligned}
    \text{ULFGW}_{r}(\mu,\nu) &\eqdef\!\!\!\! \min_{P\text{:}~\text{rk}_+(P)\leq r}\!\!\!\!\alpha |P| \langle C,P\rangle + \bar\alpha\mathcal{Q}_{A,B}(P) +\tau_1 \text{KL}(P\mathbf{1}_m|a) + \tau_2 \text{KL}(P^T\mathbf{1}_n|b)\,,
\end{aligned}
\end{equation}
which is expanded through the explicit factorization of $P$, noticing that $|P| = |g| := \|g\|_1$:
\begin{equation}
\label{eq-reformulated-ufgw}
\begin{aligned}
    \text{ULFGW}_{r}(\mu,\nu) &\eqdef \min_{(Q,R,g)\in\Pi_r}\alpha|g|  \mathcal{L}_C(Q,R,g) + \bar\alpha\mathcal{Q}_{A,B}(Q,R,g) + \mathcal{G}_{a,b}(Q,R,g)
    \end{aligned}
\end{equation}
 Then by linearizing again $\mathcal{H}:(Q,R,g)\to\alpha|g|  \mathcal{L}_C(Q,R,g) + \bar\alpha\mathcal{Q}_{A,B}(Q,R,g)$ with an added KL penalty and leaving $\mathcal{G}_{a,b}$ unchanged, we obtain at each iteration, the same optimization problem as in~\eqref{eq-barycenter-ulgw} where the kernel $\bm{\xi}_k$ is now defined as
\begin{align*}
    \bm{\xi}_k &\eqdef (\xi_{k}^{(1)},\xi_{k}^{(2)},\xi_{k}^{(3)}),\\
    \xi_{k}^{(1)} &\eqdef Q_k \odot \exp(-\gamma_k\nabla_Q\mathcal{H}_k),~\xi_{k}^{(2)} \eqdef R_k \odot \exp(-\gamma_k\nabla_Q\mathcal{H}_k),~\xi_{k}^{(3)} \eqdef g_k \odot \exp(-\gamma_k\nabla_g\mathcal{H}_k)\\
    \nabla_Q\mathcal{H}_k&\eqdef \alpha |g_k|CR_k\Diag(1/g_k) + \bar\alpha\left(2 Q_k\mathbf{1}_r\mathbf{1}_r^T + 4\Dista P_k\Distb R_k\Diag(1/g_k) \right)\\
    \nabla_R\mathcal{H}_k&\eqdef \alpha |g_k|C^T Q_k\Diag(1/g_k) + \bar\alpha\left(2 R_k\mathbf{1}_r\mathbf{1}_r^T + 4\Distb P_k^T\Dista Q_k\Diag(1/g_k) \right)\\
    \nabla_g\mathcal{H}_k&\eqdef\alpha\left( \langle C,P_k\rangle\mathbf{1}_r -|g_k|\omega_k^{\text{lin}}/g_k^2 \right) - 4\bar\alpha\omega_k^{\text{quad}}/g_k^2\\
    [\omega_k^{\text{lin}}]_i &\eqdef [Q_k^TCR_k]_{i,i},~~[\omega_k^{\text{quad}}]_i \eqdef [Q_k^T\Dista P_k\Distb R_k]_{i,i}~~\forall~ i\in\{1,\dots,r\}\; .
\end{align*}
See Algorithm~\ref{alg-prox-ulfw} for our implementatoin for the unbalanced and low-rank fused GW problem. 
\begin{algorithm}[H]
\SetAlgoLined
\textbf{Inputs:} $\Dista, \Distb, C, a, b, r, t, \gamma_0, \tau_1,\tau_2,\delta$\\
$Q,R,g \gets \text{Initialization as proposed in~\citep{scetbon2022lot}}$\\
\Repeat{$\Delta((Q,R,g),(\tilde Q,\tilde R,\tilde g),\gamma)<\delta$}{
    $\tilde{Q}=Q,~~\tilde{R}=R,~~\tilde{g}=g,\\
    \nabla_Q = \alpha |g|CR\Diag(1/g) + \bar\alpha\left(2 Q\mathbf{1}_r\mathbf{1}_r^T + 4\Dista Q\Diag(1/g)R^T\Distb R\Diag(1/g) \right),\\
    \nabla_R = \alpha |g|C^T Q\Diag(1/g) + \bar\alpha\left(2 R\mathbf{1}_r\mathbf{1}_r^T + 4\Distb R\Diag(1/g)Q^T\Dista Q\Diag(1/g) \right),\\
    \omega^{\text{lin}} \gets \mathcal{D}(Q^TCR),~~ \omega^{\text{quad}} \gets \mathcal{D}(Q^T\Dista Q\Diag(1/g)R^T\Distb R)\\
    \nabla_g =\alpha\left( \langle C,Q\Diag(1/g)R^T\rangle\mathbf{1}_r -|g_k|\omega^{\text{lin}}/g^2 \right) - 4\bar\alpha\omega^{\text{quad}}/g^2,\\
    \gamma \gets \gamma_0 / \max(\|\nabla_Q\|_{\infty}^2,\|\nabla_R\|_{\infty}^2, \|\nabla_g\|_{\infty}^2),\\
    \xi^{(1)}\gets Q \odot \exp(-\gamma\nabla_Q),~ \xi^{(2)}\gets R\odot \exp(-\gamma\nabla_R),~\xi^{(3)}\gets g\odot \exp(-\gamma_k\nabla_g),\\
    Q,R,g\gets \text{ULR-TI-Dykstra}(a,b,\bm{\xi},\gamma,\tau_1,\tau_2,\delta)~(\text{Alg.~\ref{alg-dykstra-ti-ulot})}$
      }
\textbf{Result:} $Q,R,g$
\caption{$\text{ULFGW}(\Dista,\Distb,a,b,r,\gamma_0,\tau_1,\tau_2,\delta)$ \label{alg-prox-ulfw}}
\end{algorithm}
\begin{rmq}
Note again that here, we have in general a quadratic complexity both in time and memory, and as soon as we are provided a low-rank approximation of the matrices $C,\Dista,\Distb$, our proposed algorithm scales linearly with respect to the number of points $n$ and $m$. 
\end{rmq}

	\section{Experiments}\label{sec:exp}
Our goal in \textbf{Exp. 1} is to compare unbalanced and low-rank (ULR) solvers to balanced and low-rank (LR) counterparts, in \textbf{Exp. 2}, compare ULR solvers to entropic (E) counterparts, and in \textbf{Exp. 3}, compare our ULR solvers to \citep{Thual2022unbfused}, which can learn a sparse transport map, in the unbalanced FGW setting.

\textbf{Datasets.} We run the experiments on two real world datasets, described in \ref{app:dataproc}, that are large enough to showcase our solvers. In particular, they consist of both a shared feature space, used to compute the costs matrices for the linear term in the OT and FGW settings, as well as geometries specific to each source $s$ and target $t$ data, that are used to compute the costs matrices for the quadratic term in the GW and FGW settings. We leverage mouse brain STARmap spatial transcriptomics data from \citep{Shi2022starmapbrain} for \textbf{Exp. 1} and \textbf{Exp. 2}. For \textbf{Exp. 3} we use data from the Individual Brain Charting dataset \citep{pinho2018fmribrain}, to recapitulate the settings of \citep{Thual2022unbfused}.

\textbf{Metrics.} Following \cite{Klein2023.05.11.540374}, we evaluate maps by focusing on the two following metrics: (i) \textbf{pearson correlation $\rho$} computed between the source $s$ feature matrix $F^{s}$ and the barycentric projection of the target $t$ to the source scaled by the target marginals $b^{t}$:  $T^{T}_{t \rightarrow s}\left(F^{t}\right.\frac{1}{b^{t}})$; (ii) \textbf{F1 score} computed between the original source $s$ labels $l^{s}$ and the inferred source labels, computed by taking the $\operatorname*{argmax}_j B_{i,j}$ of the barycentric projection of the target $t$ one hot encoded labels $L^{t}$, scaled by the target marginal $b^{t}$, to the source $T^{T}_{t \rightarrow s}\left(L^{t}\right.\frac{1}{b^{t}})$.

\subsection{Experiment 1: ULOT vs. LOT on gene expression / cell type annotation mapping}\label{exp:exp1}

\begin{wraptable}{R}{7cm}
\vspace*{-4mm}
\centering
\scalebox{0.6}{\begin{tabular}{lccccccc}
\toprule
\textbf{solver} & \textbf{mass pct} & \textbf{val $\rho$} & \textbf{test $\rho$} & \textbf{F1 macro} & \textbf{F1 micro} & \textbf{F1 weighted} \\
\midrule
LOT & 1.000 & 0.282 & 0.386 & 0.210 & 0.411 & 0.360 \\
ULOT & 0.889 & 0.301 & 0.409 & 0.200 & 0.425 & 0.363 \\
\midrule
LGW & 1.000 & 0.227 & 0.288 & 0.487 & 0.716 & 0.692 \\
ULGW & 1.001 & 0.222 & 0.287 & 0.463 & 0.701 & 0.665 \\
\midrule
LFGW & 1.000 & 0.365 & 0.443 & 0.576 & 0.720 & 0.714 \\
ULFGW & 0.443 & \textbf{0.379} & \textbf{0.463} & \textbf{0.582} &\textbf{ 0.733} & \textbf{0.724} \\
\bottomrule
\end{tabular}}
\vspace*{-2mm}
\caption{Results for spatial transcriptomics dataset (brain coronal section from \cite{Shi2022starmapbrain}).}
\label{exp:table-exp1}
\vspace*{-3mm}
\end{wraptable}

\begin{figure}
\centering
\begin{subfigure}[b]{.48\textwidth}
    \centering
    \includegraphics[width=.98\textwidth]{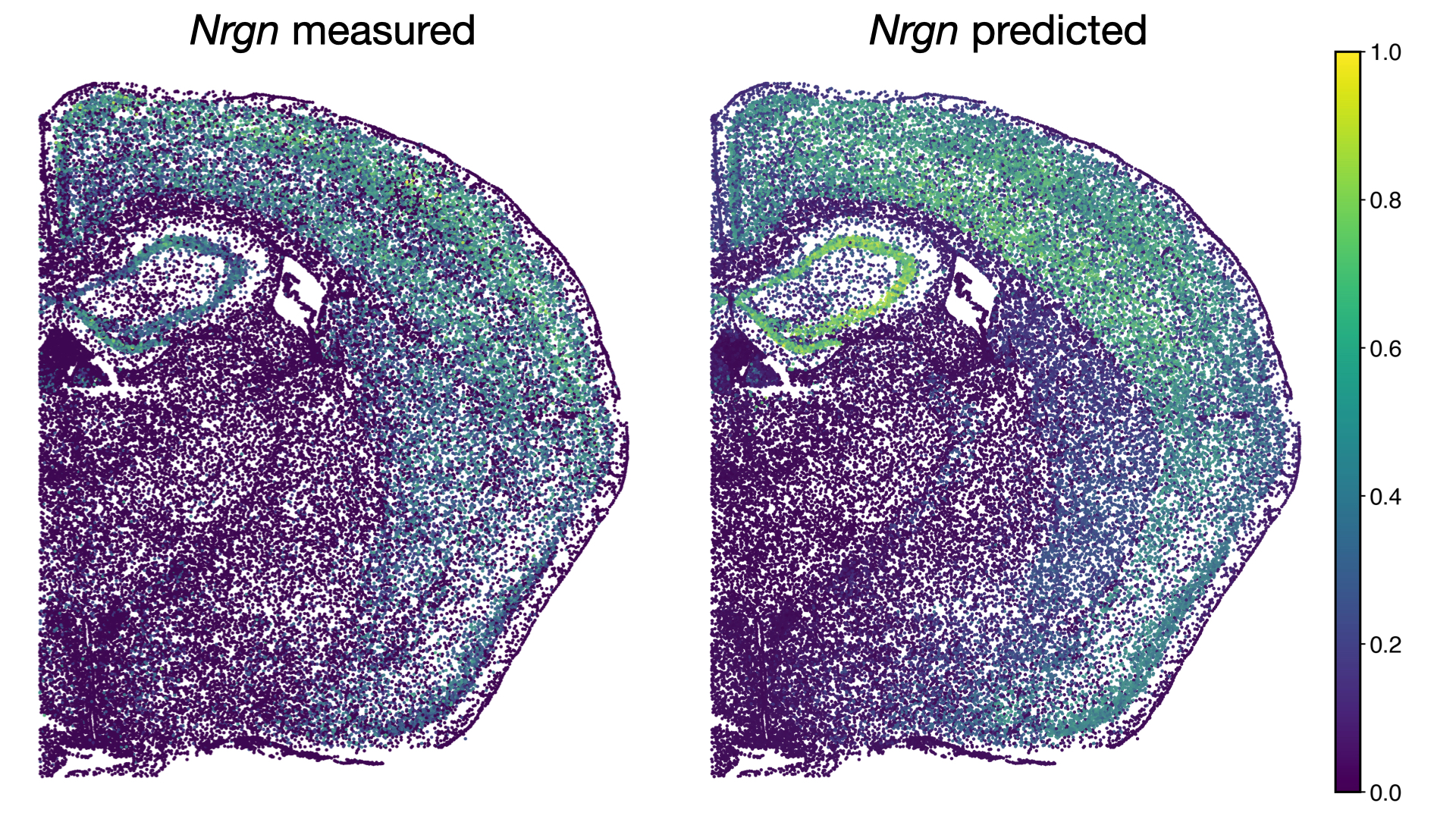}
    \caption{Visualization of measured and predicted gene expression of \textit{Nrgn}.}
    \label{exp:figure-genes-nrgn}
\end{subfigure}%
\hfill
\begin{subfigure}[b]{.48\textwidth}
    \centering
    \includegraphics[width=.98\textwidth]{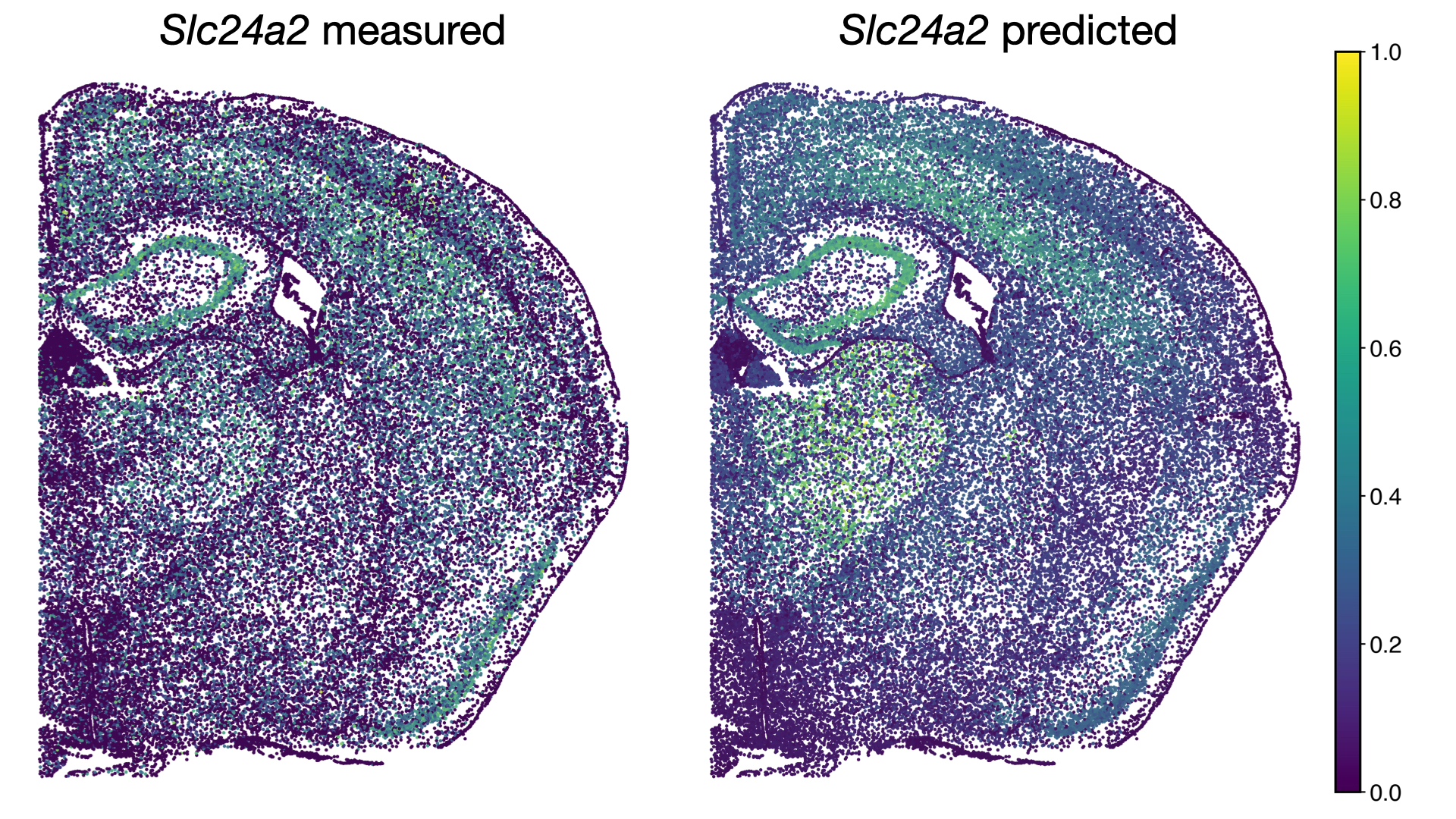}
    \caption{Visualization of measured and predicted gene expression of \textit{Slc24a2}.}
    \label{exp:figure-genes-slc24a2}
\end{subfigure}%
\caption{Spatial visualization of the two mouse brain sections used in \textbf{Exp. 1}}
\label{exp:figure-genes}
\end{figure}

Here, we evaluate the accuracy of ULOT solvers for a large-scale spatial transcriptomics task, using gene expression mapping and cell type annotation. We compare it to the balanced LR alternative using the Pearson correlation $\rho$ as described in the metrics section. We leverage two coronal sections of the mouse brain profiled by STARmap spatial transcriptomics by \citep{Shi2022starmapbrain}. They consist of $n\approx 40,000$ cells in both the source and target brain section. Each cell is described by 1000 gene features, in addition to 2D spatial coordinates. As a result $A,B$ are $\approx 40k\times 40k$, and the fused term $C$ is a squared-Euclidean distance matrix on 30D PCA space computed on the gene expression space. We selected 10 marker genes for the validation and test sets from the \textit{HPF\_CA} cluster.

\begin{wrapfigure}{r}{0.5\textwidth}
\vspace*{-6mm}
\centering
\includegraphics[width=.45\textwidth]{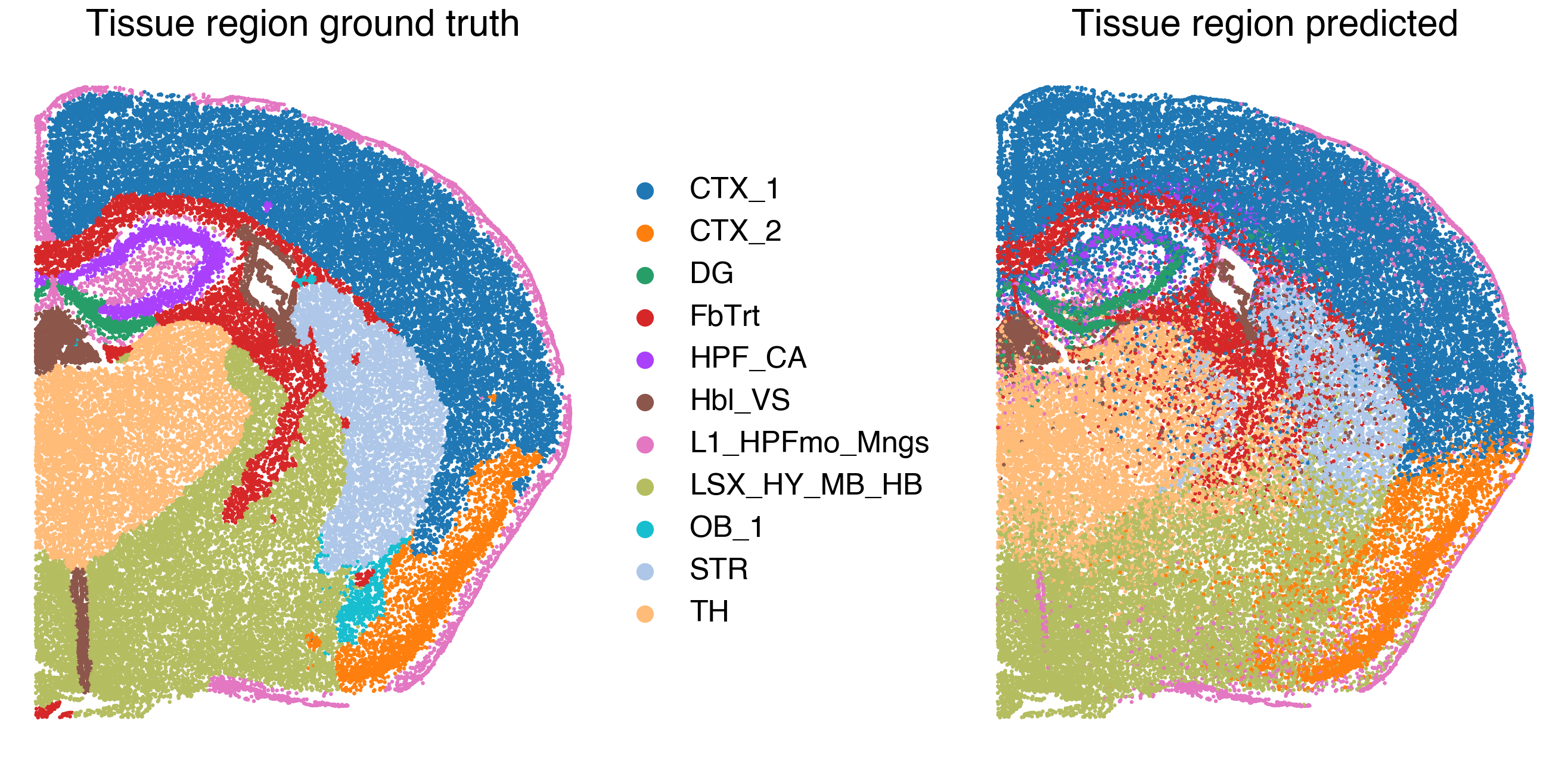}
\caption{Visualization of measured and predicted tissue regions in the mouse brain in \textbf{Exp. 1}}
\label{exp:figure-celltypes}
\vspace*{-2mm}
\end{wrapfigure}

We run an extensive grid search as reported in \ref{app:exp-grid}, we pick the best hyperparameters combination using performance on the $10$ validation genes as a criterion, and we report that metric on the other genes in Table~\ref{exp:table-exp1}, as well as qualitative results in Figure~\ref{exp:figure-genes} and Figure~\ref{exp:figure-celltypes}. Clearly, ULFGW is the best performing solver across all metrics. Interestingly, the ULOT does not consistently outperforms its balanced version, and unbalancedness seems to hurt performance for the LGW solvers. Nevertheless, both solvers display inconsistent performance across metrics, whereas the ULFGW and LFGW are consistently superior to the rest of the solvers. These results highlight how the flexibility given by the FGW formulation to leverage common and disparate geometries, paired with the unbalancedness relaxation, can provide state of the art algorithms for matching problems in large-scale, real world biological problems.

\subsection{Experiment 2: ULOT vs. UEOT}\label{exp:exp2}
In this experiment, we evaluate the performance of ULOT solvers to the unbalanced entropic alternative (UEOT). We use the same datasets as \ref{exp:exp1}, but pick a smaller subset (Olfactory bulb), to avoid OOM errors for entropic UGW solvers, that cannot handle the $40k$ sizes considered previously (see \ref{app:dataproc}).

\begin{wraptable}{R}{6cm}
\centering
\scalebox{0.6}{\begin{tabular}{lccccccc}
\toprule
\textbf{solver} & \textbf{mass} & \textbf{val $\rho$} & \textbf{test $\rho$} & \textbf{F1-mac} & \textbf{F1-mic} & \textbf{F1-wei} \\
\midrule
UEOT & 1.012 & 0.368 & 0.479 & 0.511 & 0.763 & 0.751 \\
LOT & 1.000 & 0.335 & 0.440 & 0.511 & 0.760 & 0.751 \\
ULOT & 0.998 & 0.356 & 0.461 & 0.518 & 0.770 & 0.762 \\
\midrule
UEFGW & 1.015 & 0.343 & 0.475 & \textbf{0.564} & \textbf{0.839} &\textbf{ 0.831} \\
LFGW & 1.000 & 0.348 & 0.453 & 0.512 & 0.762 & 0.753 \\
ULFGW & 0.339 & \textbf{0.368} & \textbf{0.491} & 0.556 & 0.826 & 0.818 \\
\bottomrule
\end{tabular}}
\vspace*{-1mm}
\caption{Results for spatial transcriptomics dataset (Olfactory bulb section from \cite{Shi2022starmapbrain}).}
\label{exp:table-exp2}  
\end{wraptable}

They consist of $n\approx 20,000$ cells in the source and $\approx 15,000$ cells in the target sections, and 1000 genes. Similar to \textbf{Exp.1}, the fused term $C$ is a squared-Euclidean distance matrix on 30D PCA space computed on the gene expression space. As done in the previous experiment, we select 10 marker genes for the validation and 10 genes for the test set, from cluster \textit{OB\_1}. We run an extensive grid search as described for \textbf{Exp.2} in \ref{exp:exp1} and \ref{app:exp-grid}. In Table~\ref{exp:table-exp2}, we see that ULFGW outperforms entropic solvers w.r.t. $\rho$, but is worse when considering the F1 scores. On the other hand, ULFGW confirms its superiority compared to the balanced alternative LFGW. Taken together, these results suggest that while unbalanced LR solvers are on par with unbalanced entropic solvers in terms of performance on small data regimes, they unlock the applications of unbalanced OT to large scale datasets.

\subsection{Experiment 3: ULOT to align brain meshes}\label{exp:exp2}
\citet{Thual2022unbfused} proposed a novel formulation for the unbalanced FGW problem. Their proposal is showcased to align brain anatomies, and their functional signal (FUGW). We compare it to our ULFGW solver, using the same experimental setting, see Table~\ref{exp:table-exp3}.

\begin{wraptable}{R}{5cm}
\centering
\scalebox{0.8}{\begin{tabular}{lccc}
\toprule
\textbf{solver} & \textbf{mass} & \textbf{val $\rho$} & \textbf{test $\rho$} \\
\midrule
FUGW-sparse & 0.999 & 0.492 & 0.472 \\
LFGW & 1.000 & 0.513 & \textbf{0.663} \\
ULFGW & 0.981 & \textbf{0.533} & 0.643 \\
\bottomrule
\end{tabular}}
\vspace*{-1mm}
\caption{Results on the brain anatomy with functional signal data from \cite{pinho2018fmribrain} in \textbf{Exp.3}.}
\label{exp:table-exp3}
\end{wraptable}

\begin{figure}
\centering
\includegraphics[width=.98\textwidth]{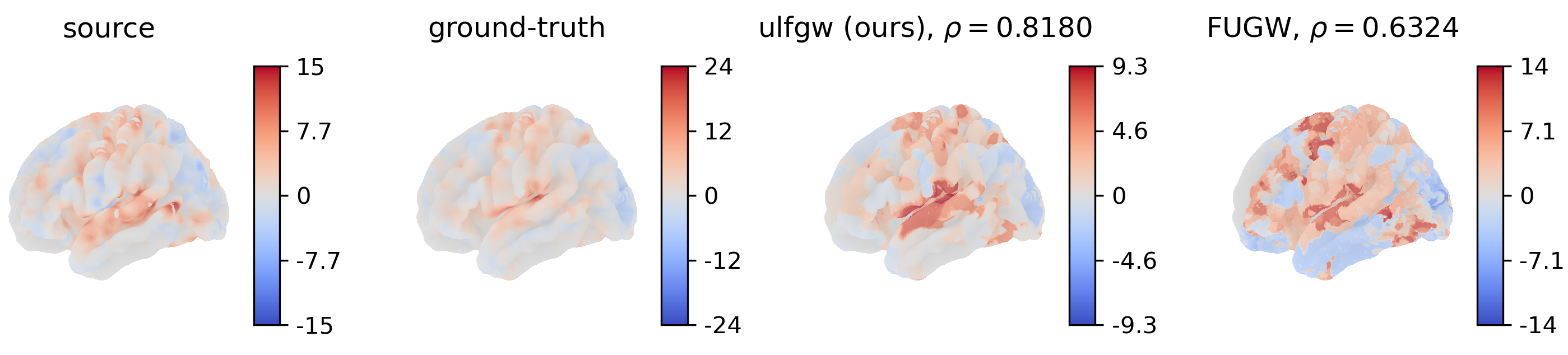}
\caption{Visualization of measured and predicted \textit{right auditory click} contrast map in \textbf{Exp.3}.}
\label{exp:figure-brain-right}
\end{figure}

\subsection*{Conclusion}
Recent practical successes of OT methods to natural sciences have demonstrated the relevance of OT to their analysis pipelines, but have also shown, repeatedly, that a certain degree of freedom to depart from the rigid assumption of mass conservation is needed in practice. On the other hand, and across the same range of applications, low-rank approaches can hold the promise of scaling OT methods to relevant sample sizes for natural sciences. This paper merges these two strains and demonstrate the practical relevance of these novel algorithms.

\clearpage
\newpage

\bibliography{biblio}
\bibliographystyle{plainnat}

\newpage
\appendix


\section*{Appendix}
\label{appendix}

\section{Algorithms}

\begin{algorithm}[H]
\SetAlgoLined
\textbf{Inputs:} $a, b,\bm{\xi}=(\xi^{(1)}, \xi^{(2)},\xi^{(3)}),\gamma,\tau_1,\tau_2,\delta$\\
$v_1=v_2=\mathbf{1}_r$, $u_1=\mathbf{1}_n$, $u_2=\mathbf{1}_m$\\
\Repeat{$\frac{1}{\gamma}\max(\Vert \log(u_i/\tilde{u}_i)\|_{\infty}, \|\log(v_i/\tilde{v}_i)\|_{\infty})<\delta$}{
    $\tilde{v}_1=v_1,~\tilde{v}_2=v_2, \tilde{u}_1 = u_1, \tilde{u}_2 = u_2\\
    u_{1} = \left(\frac{a}{\xi^{(1)}v_{1}}\right)^{\frac{\tau_1}{\tau_1+1/\gamma}}, \quad u_{2} = \left(\frac{b}{\xi^{(2)}v_{2}}\right)^{\frac{\tau_2}{\tau_2+1/\gamma}},\\
   g = \left(\xi^{(3)}\odot  (\xi^{(1)})^T u_1 \odot (\xi^{(2)})^T u_2\right)^{1/3},~v_1 = \frac{g}{(\xi^{(1)})^T u_1},~ v_2 = \frac{g}{(\xi^{(2)})^T u_2}$
  }
\textbf{Result:} $\Diag(u_1)\xi^{(1)}_k\Diag(v_1),~~\Diag(u_2)\xi^{(2)}_k\Diag(v_2),~~g$
\caption{$\text{ULR-Dykstra}(a,b,\bm{\xi},\gamma,\tau_1,\tau_2,\delta)$~\label{alg-dykstra-ulot}}
\end{algorithm}

\section{Experiments}
\subsection{Datasets and preprocessing}
\label{app:dataproc}
\begin{figure}[!h]
\centering
\begin{subfigure}[b]{.48\textwidth}
    \centering
    \includegraphics[width=.98\textwidth]{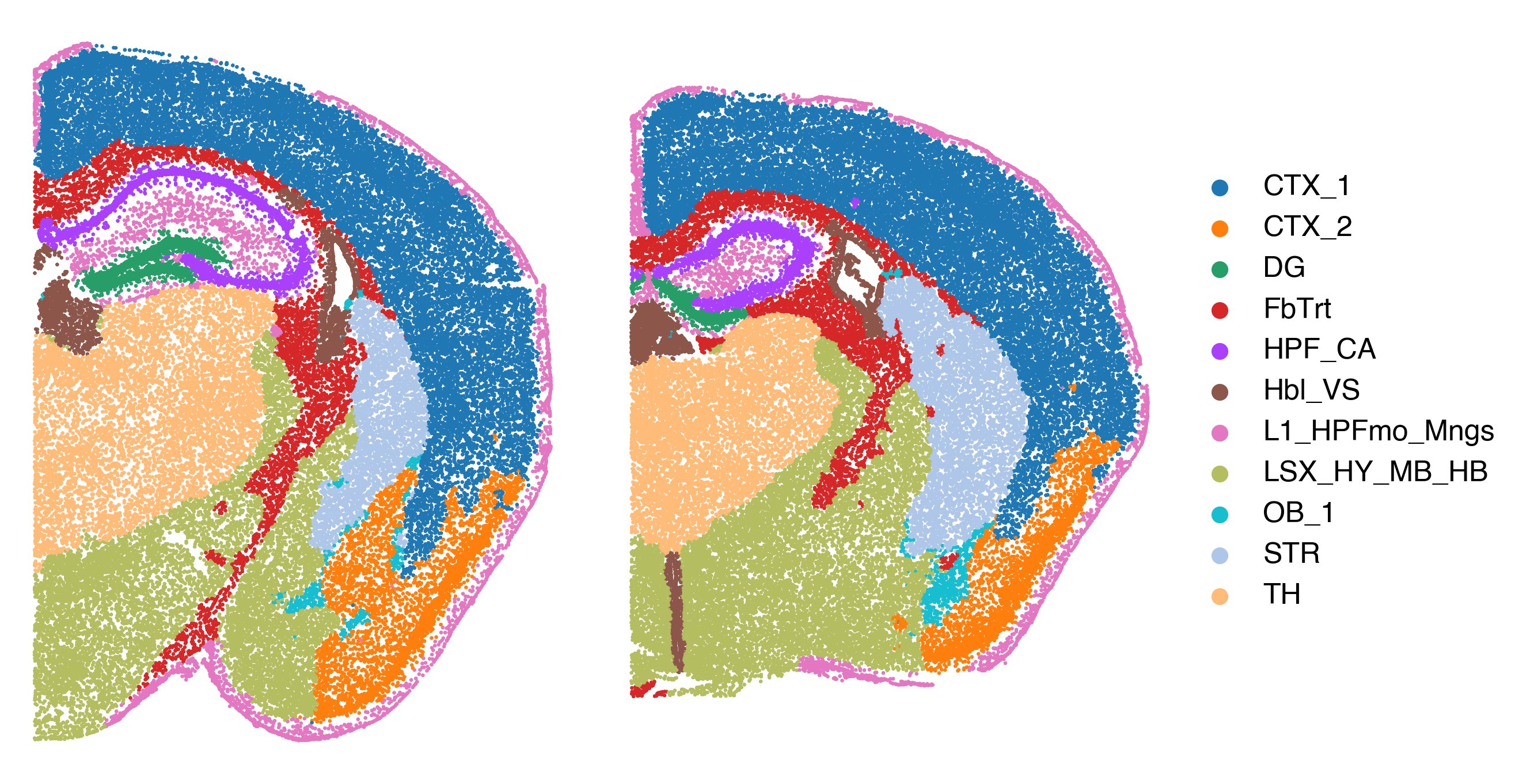}
    \caption{Visualization of the brain coronal sections used in \textbf{Exp.1}.}
    \label{appfig:brain-coronal-spatial}
\end{subfigure}%
\hfill
\begin{subfigure}[b]{.48\textwidth}
    \centering
    \includegraphics[width=.98\textwidth]{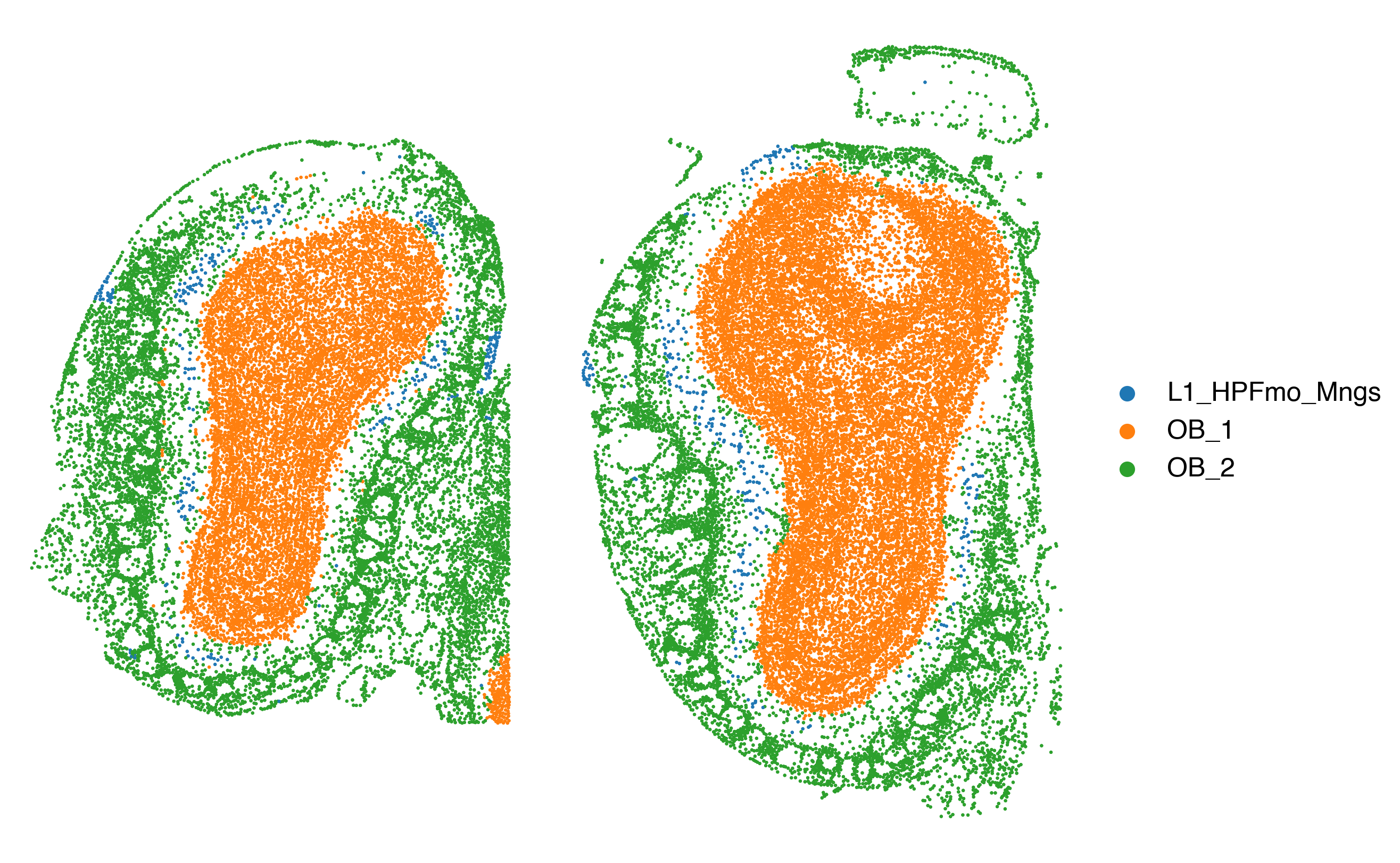}
    \caption{Visualization of the OB sections used in \textbf{Exp.2}.}
    \label{appfig:ob-coronal-spatial}
\end{subfigure}%
\caption{Spatial visualization of the two mouse brain sections used in \textbf{Exp.1}.}
\label{appfig:coronal-spatial}
\end{figure}

\begin{figure}
\centering
\begin{subfigure}[b]{0.48\linewidth}
    \centering
    \includegraphics[width=.98\textwidth]{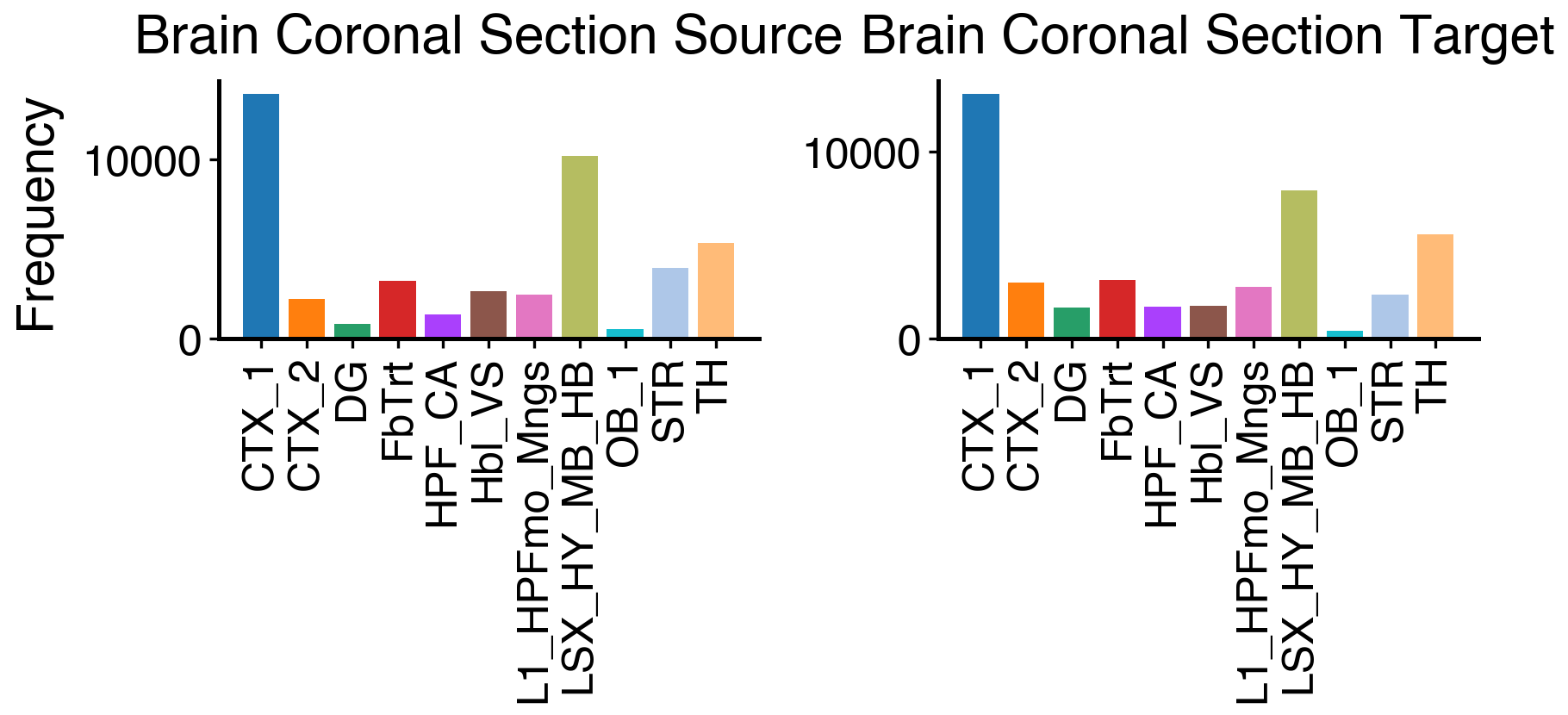}
    \caption{Visualization of cell type frequencies for the brain coronal sections used in \textbf{Exp.1}.}
    \label{appfig:brain-coronal-prop}
\end{subfigure}%
\hfill
\begin{subfigure}[b]{0.48\linewidth}
    \centering
    \includegraphics[width=.98\textwidth]{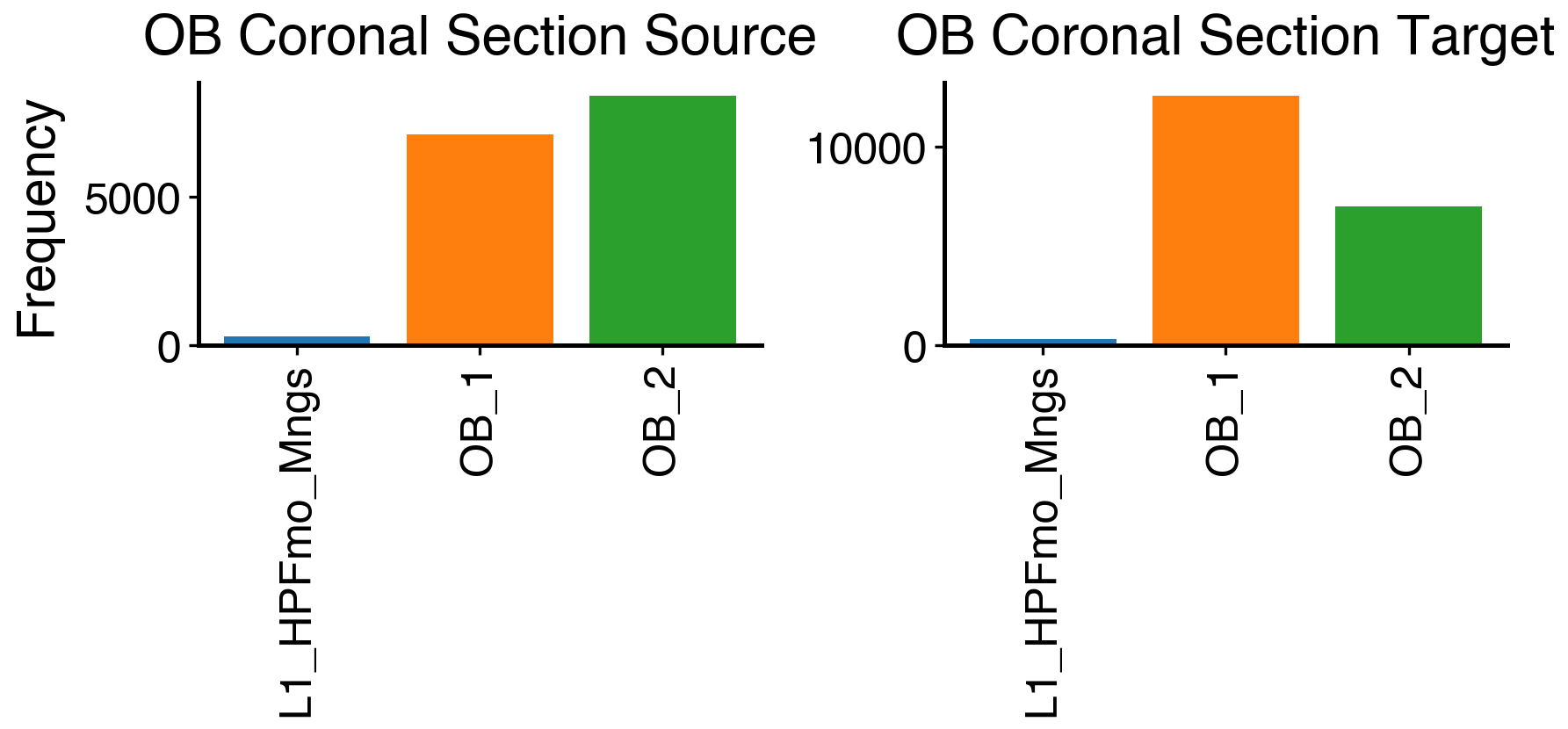}
    \caption{Visualization of cell type frequencies for the OB sections used in \textbf{Exp.2}.}
    \label{appfig:ob-coronal-prop}
\end{subfigure}%
\caption{Cell type frequencies of the datasets used in \textbf{Exp.1} and \textbf{Exp.2}.}
\label{appfig:proportions}
\end{figure}

We downloaded the two publicly available datasets from the respective publications:
\begin{itemize}
    \item STARmap mouse brain sections from \citep{Shi2022starmapbrain}
    \item Brain mesh anatomy and functional signal from \citep{pinho2018fmribrain}
\end{itemize}
We reprocessed the datasets using standard tools from the SCANPY pipeline \citep{Wolf2018scanpy}. Specifically, we log-normalized gene expression of all genes present in dataset. We selected two brain coronal sections for \textbf{Exp.1} and two Coronal Olfactory Bulb (OB) sections for \textbf{Exp.2}, from the STARmap dataset. For \textbf{Exp.3}, we used the meshes together with their functional signal of the brains to recapitulate \textbf{Exp.1} in \citep{Thual2022unbfused}. A visualization of the STARmap dataset for the two subsets used in \textbf{Exp.1} and \textbf{Exp.2} can be seen in Figure~\ref{appfig:coronal-spatial} and an overview of the cell type proportions present in each of the section pairs can be see in Figure~\ref{appfig:proportions}. These visualization highlight the differences in terms of spatial organization and cell type proportions of the brain sections used in the experiment.

\subsection{Experimental settings}
\label{app:exp-grid}
For the brain signal alignment benchmark in \textbf{Exp.3}, we use the \textit{fsaverage7} mesh containing 160k vertices. We compute an approximation of the geodesic distances with Landmark Multi-Dimensional Scaling \cite{de2004sparse} using 2048 points in 30-dimensional embedding space. For \textbf{fugw-sparse}, we compute the coupling in 2 stages:

\begin{enumerate}
    \item Similarly as in \cite{Thual2022unbfused}, we subsample the mesh to 10\% of the points using Ward's algorithm and compute the coarse optimal transport coupling.
    \item We then use this coarse coupling to define a sparsity mask on the full mesh by selecting for each source (target) vertex the most coupled target (source) vertex and its neighbors within $\frac{4}{max\_distance}$ radius using the approximation of the geodesic distances. This mask is then used to compute the fine-grained sparse coupling.
\end{enumerate}

For all experiments, we ran the grid search as defined by \ref{tab:grid} and selected the best set of hyperparameters based on the validation correlation. We report results of top performing hyperparameters for the evaluated algorithms in Table~\ref{apptable:table-exp1} for \textbf{Exp.1}, Table~\ref{apptable:table-exp2} for \textbf{Exp.2} and Table~\ref{apptable:table-exp3} for \textbf{Exp.3}

\begin{table}[!h]
\centering
\begin{tabular}{ll}
\toprule
 & \textbf{values} \\
\midrule
\textbf{rank} & {10, 50, 100} \\
\textbf{reg (ours)} & {0.0, 0.001, 0.01} \\
\textbf{reg (fugw-sparse)} & {0.0001, 0.001, 0.01} \\
\textbf{tau1} & {0.1, 1.0, 100.0} \\
\textbf{tau2} & {0.1, 1.0, 100.0} \\
\bottomrule
\end{tabular}
\caption{Hyperparameters considered in our grid-search.}
\label{tab:grid}
\end{table}

\begin{table}[!h]
\centering
\scalebox{0.9}{\begin{tabular}{lrrrrrrrrrrr}
\toprule
\textbf{solver} & \textbf{rank} & \textbf{tau1} & \textbf{tau2} & \textbf{temp} & \textbf{reg} & \textbf{mass} & \textbf{val $\rho$} & \textbf{test $\rho$} & \textbf{F1-mac} & \textbf{F1-mic} & \textbf{F1-wei} \\
\midrule
lot & 10 & - & - & 0.200 & 0.010 & 1.000 & 0.282 & 0.386 & 0.210 & 0.411 & 0.360 \\
ulot & 10 & 1.000 & 1.000 & 0.200 & 0.010 & 0.889 & 0.301 & 0.409 & 0.200 & 0.425 & 0.363 \\
lgw & 100 & - & - & 0.200 & 0.001 & 1.000 & 0.227 & 0.288 & 0.487 & 0.716 & 0.692 \\
ulgw & 100 & 100.000 & 100.000 & 0.200 & 0.010 & 1.001 & 0.222 & 0.287 & 0.463 & 0.701 & 0.665 \\
lfgw & 50 & - & - & 0.400 & 0.010 & 1.000 & 0.365 & 0.443 & 0.576 & 0.720 & 0.714 \\
ulfgw & 100 & 0.100 & 0.100 & 0.400 & 0.001 & 0.443 & 0.379 & 0.463 & 0.582 & 0.733 & 0.724 \\
\bottomrule
\end{tabular}}
\vspace*{3mm}
\caption{Results on the large spatial transcriptomics dataset (brain coronal section from \citep{Shi2022starmapbrain}).}
\label{apptable:table-exp1}
\end{table}

\begin{table}[!h]
\centering
\scalebox{0.9}{\begin{tabular}{lrrrrrrrrrrr}
\toprule
\textbf{solver} & \textbf{rank} & \textbf{tau1} & \textbf{tau2} & \textbf{temp} & \textbf{reg} & \textbf{mass} & \textbf{val $\rho$} & \textbf{test $\rho$} & \textbf{F1-mac} & \textbf{F1-mic} & \textbf{F1-wei} \\
\midrule
uot & - & 0.909 & 0.999 & 0.400 & 0.100 & 1.012 & 0.368 & 0.479 & 0.511 & 0.763 & 0.751 \\
lot & 10 & - & - & 0.200 & 0.010 & 1.000 & 0.335 & 0.440 & 0.511 & 0.760 & 0.751 \\
ulot & 10 & 1.000 & 100.000 & 0.200 & 0.010 & 0.998 & 0.356 & 0.461 & 0.518 & 0.770 & 0.762 \\
ufgw & - & 0.500 & 0.999 & 0.600 & 0.100 & 1.015 & 0.343 & 0.475 & 0.564 & 0.839 & 0.831 \\
lfgw & 10 & - & - & 0.600 & 0.010 & 1.000 & 0.348 & 0.453 & 0.512 & 0.762 & 0.753 \\
ulfgw & 10 & 0.100 & 0.100 & 0.600 & 0.001 & 0.339 & 0.368 & 0.491 & 0.556 & 0.826 & 0.818 \\
\bottomrule
\end{tabular}}

\vspace*{3mm}
\caption{Results on the small subset STARmap dataset (OB section from \citep{Shi2022starmapbrain}).}
\label{apptable:table-exp2}
\end{table}

\begin{table}[!h]
\centering
\begin{tabular}{lrrrrrrrr}
\toprule
\textbf{solver} & \textbf{rank} & \textbf{tau1} & \textbf{tau2} & \textbf{reg} & \textbf{reg} & \textbf{mass} & \textbf{val $\rho$} & \textbf{test $\rho$} \\
\midrule
\textbf{fugw-sparse} & - & 1.000 & 0.100 & 0.200 & 0.01 & 0.999 & 0.492 & 0.472 \\
\textbf{lfgw} & 100 & - & - & 0.600 & 0.000 & 1.000 & 0.513 & 0.663 \\
\textbf{ulfgw} & 100 & 1.000 & 0.100 & 0.600 & 0.001 & 0.981 & 0.533 & 0.643 \\
\bottomrule
\end{tabular}

\vspace*{3mm}
\caption{Results on the brain anatomy and functional signal from \citep{pinho2018fmribrain}).}
\label{apptable:table-exp3}
\end{table}

\begin{table}[!h]
\centering
\begin{tabular}{llrrrrr}
\toprule
 \textbf{experiment} &  & \textbf{val $\rho$} & \textbf{tst $\rho$} & \textbf{F1-mac} & \textbf{F1-mic} & \textbf{F1-wei} \\
\midrule
\multirow[t]{2}{*}{\textbf{Exp. 1}} & mean & 0.362 & 0.449 & 0.546 & 0.687 & 0.677 \\
 & std & 0.027 & 0.022 & 0.054 & 0.062 & 0.061 \\
\cline{1-7}
\multirow[t]{2}{*}{\textbf{Exp. 2}} & mean & 0.356 & 0.463 & 0.538 & 0.800 & 0.791 \\
 & std & 0.008 & 0.018 & 0.021 & 0.031 & 0.032 \\
\bottomrule
\end{tabular}
\caption{Effect of k-means initialization \citep{scetbon2022lot}. We report mean and standard deviation of \textit{test} criterion for ULFGW, with the best hyperparameter on validation data for each experiment. We use 5 initial seeds for \textbf{Exp. 1}. We observe more variability in validation performance for \textbf{Exp. 2}, and therefore start with 10 seeds, pruning the lowest performing 5 seeds.}
\label{apptable:table-init}
\end{table}

\begin{figure}[!h]
\centering
\begin{subfigure}[b]{.48\textwidth}
    \centering
    \includegraphics[width=.98\textwidth]{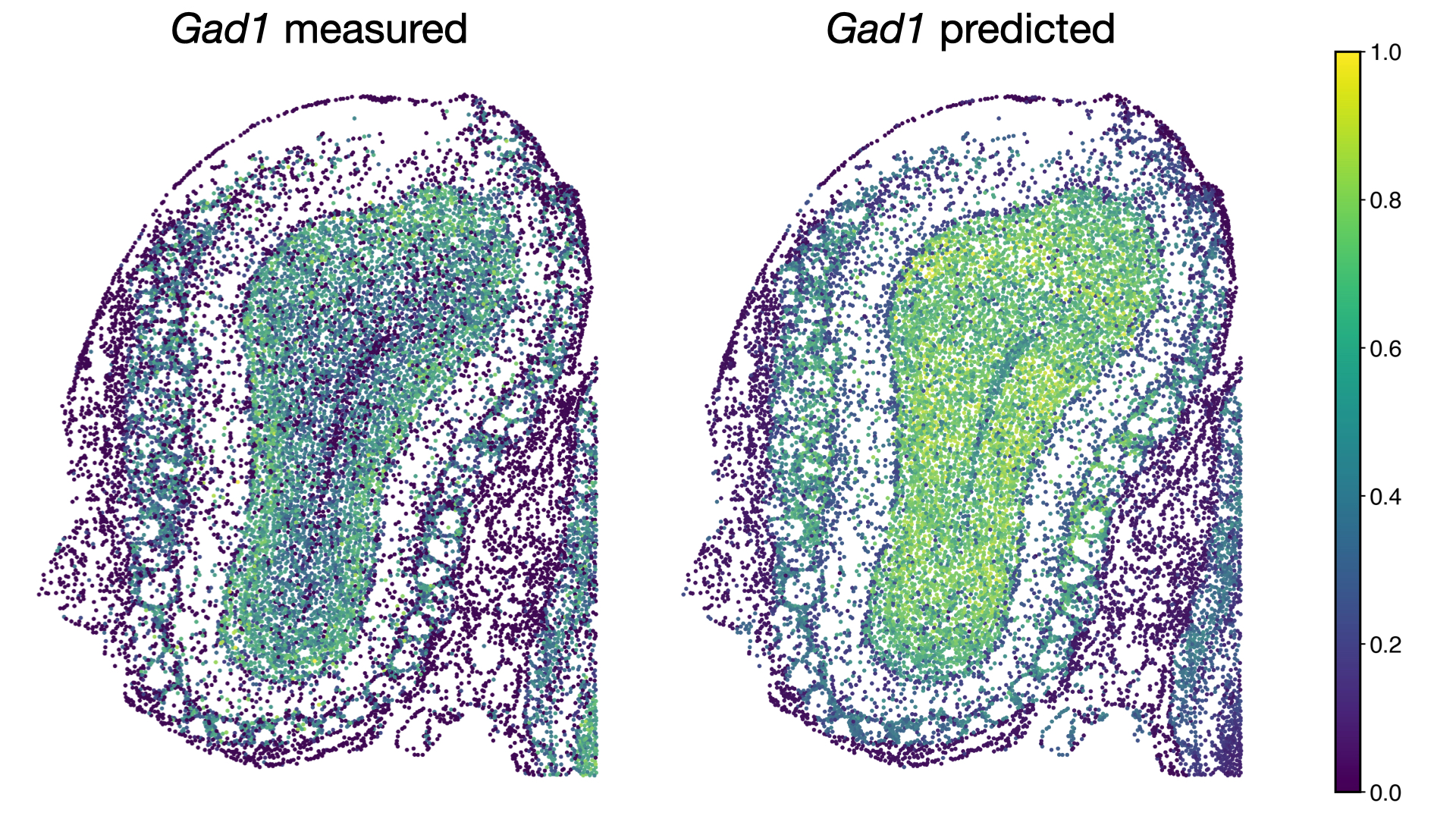}
    \caption{Visualization of measured and predicted gene expression of \textit{Gad1}.}
    \label{app:figure-genes1}
\end{subfigure}%
\hfill
\begin{subfigure}[b]{.48\textwidth}
    \centering
    \includegraphics[width=.98\textwidth]{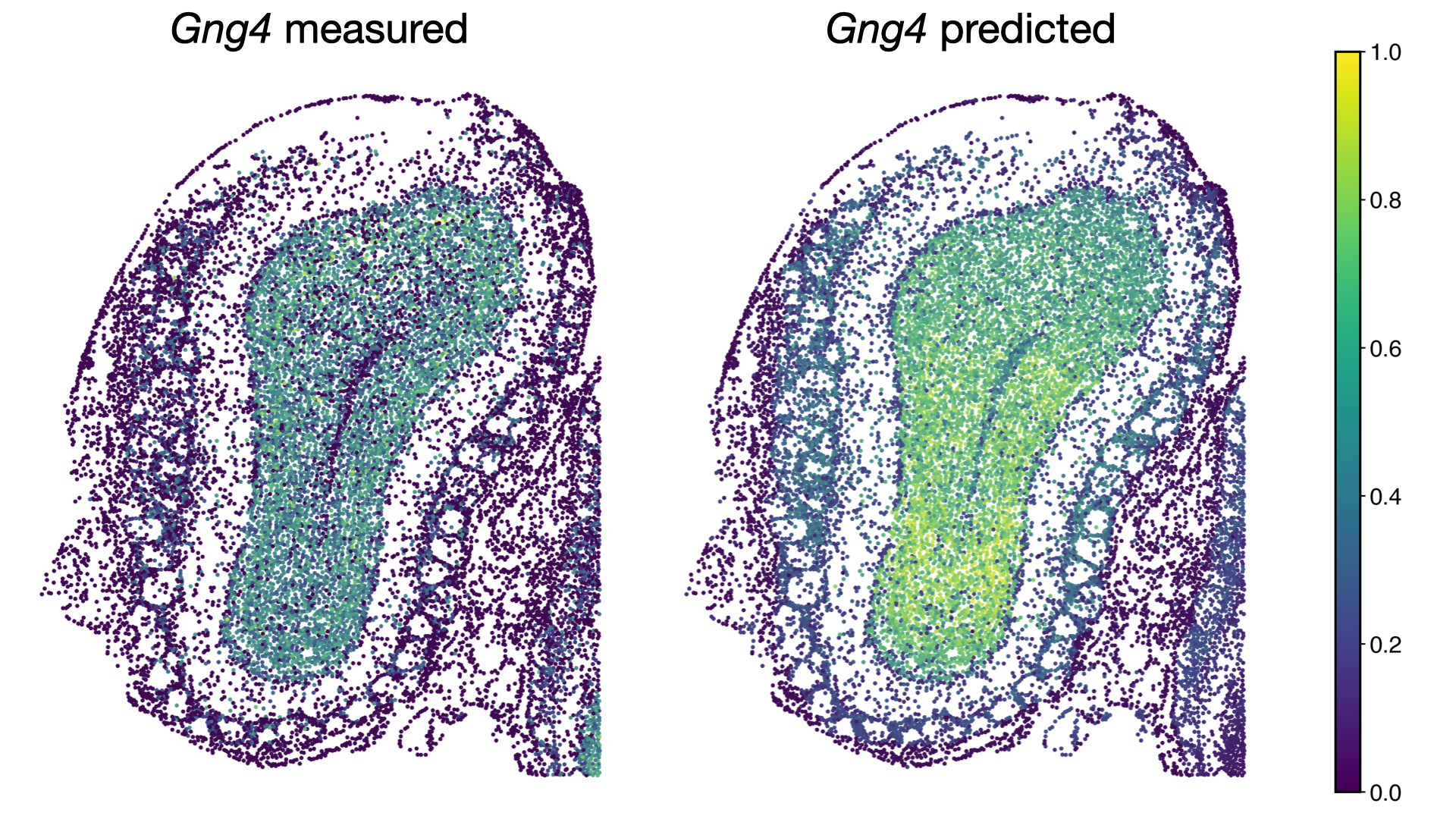}
    \caption{Visualization of measured and predicted gene expression of \textit{Gnrg4}.}
    \label{app:figure-genes1}
\end{subfigure}%
\caption{Measured and predicted gene expression for the small subset STARmap dataset (OB section from \citep{Shi2022starmapbrain}) for ULRFGW.}
\label{app:figure-genes}
\end{figure}

\section{Proofs}
\subsection{Proof of Proposition~\ref{prop-dual}}
Let $n,m\geq r\geq 1$, $\gamma>0$, $,\bm{\xi}:=(\xi^{(1)},\xi^{(2)},\xi^{(3)})$ where $\xi^{(1)}\in\mathbb{R}^{n\times r}_{+}$, $\xi^{(2)}\in\mathbb{R}^{m\times r}_{+}$ and $\xi^{(3)}\in\mathbb{R}^r_{+}$ and let us recall that $\text{KL}(\cdot,\cdot)$ is the generalized Kullback-Leibler divergence defined as $\text{KL}(p|q):=\sum_{i}p_i\log(p_i/q_i) +q_i - p_i$. Then observe that
\begin{align*}
  \min_{(Q,R,g)\in\Pi_r} \frac{1}{\gamma}\left[\text{KL}(Q,\xi^{(1)})+ \text{KL}(R,\xi^{(2)})+ \text{KL}(g,\xi^{(3)})\right]+
 \tau_1 \text{KL}(Q\mathbf{1}_r|a) + \tau_2 \text{KL}(R\mathbf{1}_r|b)
\end{align*}
is a convex problem satisfying the Slater's condition and therefore strong duality holds. Therefore we have:
\begin{align*}
  &\min_{(Q,R,g)\in\Pi_r} \frac{1}{\gamma}\left[\text{KL}(Q,\xi^{(1)})+ \text{KL}(R,\xi^{(2)})+ \text{KL}(g,\xi^{(3)})\right] +
 \tau_1 \text{KL}(Q\mathbf{1}_r|a) + \tau_2 \text{KL}(R\mathbf{1}_r|b)\\
 &=\sup_{\lambda_1,\lambda_2} \min_{Q,R,g} \langle 
 \lambda_1,g - Q^\top \mathbf{1}_n\rangle + \langle \lambda_2, g-R^{\top}\mathbf{1}_m\rangle + \frac{1}{\gamma}\left[\text{KL}(Q,\xi^{(1)})+ \text{KL}(R,\xi^{(2)})+ \text{KL}(g,\xi^{(3)})\right]\\
 &+  \tau_1 \text{KL}(Q\mathbf{1}_r|a) + \tau_2 \text{KL}(R\mathbf{1}_r|b)\\
 &=\sup_{\lambda_1,\lambda_2}\min_Q \frac{1}{\gamma}\text{KL}(Q,\xi^{(1)}) +  \tau_1 \text{KL}(Q\mathbf{1}_r|a) + \langle  \lambda_1,- Q^\top \mathbf{1}_n\rangle\\
&+ \min_R \frac{1}{\gamma}\text{KL}(R,\xi^{(2)}) + \tau_2 \text{KL}(R\mathbf{1}_r|b) +\langle 
 -\lambda_2, R^\top \mathbf{1}_m\rangle
 + \min_g \frac{1}{\gamma}\text{KL}(g,\xi^{(3)}) + \langle 
 g,\lambda_1+\lambda_2\rangle.
\end{align*} 
Now consider 
$$\min_g \frac{1}{\gamma}\text{KL}(g,\xi^{(3)}) + \langle 
 g,\lambda_1+\lambda_2\rangle$$ and observe that this problem can be solved explicitly. The first-order optimality condition gives us  that $g^{*}=\exp(-\gamma(\lambda_1+\lambda_2))\odot\xi^{(3)} $ solves the problem and 
 \begin{align*}
     \min_g \frac{1}{\gamma}\text{KL}(g,\xi^{(3)}) + \langle 
 g,\lambda_1+\lambda_2\rangle = -\frac{1}{\gamma}\langle \exp(-\gamma(\lambda_1+\lambda_2)), \xi^{(3)}\rangle + \langle \xi^{(3)},\mathbf{1}\rangle .
 \end{align*}
Let us now focus on the following convex optimization problem,
\begin{align}
\label{eq-primal-ent-simple}
\min_Q \frac{1}{\gamma}\text{KL}(Q,\xi^{(1)}) +  \tau_1 \text{KL}(Q\mathbf{1}_r|a) + \langle  -\lambda_1, Q^\top \mathbf{1}_n\rangle
\end{align}
and note that it admits a unique solution due to the strict convexity of $Q\to\text{KL}(Q,\xi^{(1)})$. Then by denoting $F_{\tau,z}(s):=\tau\text{KL}(s|z)$ and $G_{\lambda}(s):=\langle s,-\lambda\rangle$, and by applying the Fenchel-Rockafellar theorem~\citep{rockafellar1970convex}, we obtain that strong duality holds, the dual problem of~\eqref{eq-primal-ent-simple} is
\begin{align*}
    \sup_{f_1,h_1} -F_{\tau_1,a}^{*}(-f_1) - G_{\lambda_1}^{*}(-h_1) - \frac{1}{\gamma}\langle \exp(\gamma(f_1+h_1)),\xi^{(1)}\rangle
\end{align*}
and that $(f_1,h_1)$ solves the dual if and only if $-f_1\in\partial F_{\tau_1,a}(Q\mathbf{1}_r)$,$-h_1\in\partial G_{\lambda_1}(Q^\top\mathbf{1}_n)$ and 
$Q=\Diag(\exp(\gamma f_1))\xi^{(1)}\Diag(\exp(\gamma h_1))$ where $Q$ is the solution of~\eqref{eq-primal-ent-simple}. Recall that here we denote for any convex set $X\in\mathbb{R}^q$ and function $f:X\to\mathbb{R}\cup\{+\infty\}$, $f^*$ its convex conjugate defined for any $y\in X^*:=\{x^*~\text{s.t.}~\sup_{x\in X}\langle x,x^*\rangle - f(x) <+\infty \}$ by $f^{*}(y):=\sup_{x\in X} \langle x,y\rangle - f(x)$ and $\partial f(x):=\{y~\text{s.t.}~ f(x')-f(x)\geq \langle y,x-x'\rangle~ \forall x'\in X\}$. Now remarks that
\begin{align*}
   G_{\lambda_1}^{*}(-h_1) = \sup_s \langle s,\lambda_1-h_1\rangle =\left\{
    \begin{array}{ll}
        +\infty & \mbox{if } \lambda_1\neq h_1\\
        0 & \mbox{otherwise .}
    \end{array}
\right.
\end{align*}
therefore $ G_{\lambda_1}^{*}$ ensures that $\lambda_1 = h_1$. Similarly we obtain that 
\begin{align}
\min_R \frac{1}{\gamma}\text{KL}(R,\xi^{(2)}) +  \tau_2 \text{KL}(r\mathbf{1}_r|b) + \langle  -\lambda_2, R^\top \mathbf{1}_m\rangle
\end{align}
is equal to its dual defined as
\begin{align*}
    \sup_{f_2,h_2} -F_{\tau_2,b}^{*}(-f_2) - G_{\lambda_2}^{*}(-h_2) - \frac{1}{\gamma}\langle \exp(\gamma(f_2+h_2)),\xi^{(2)}\rangle
\end{align*}
where again 
\begin{align*}
   G_{\lambda_2}^{*}(-h_2) =\left\{
    \begin{array}{ll}
        +\infty & \mbox{if } \lambda_2\neq h_2\\
        0 & \mbox{otherwise}
    \end{array}
\right.
\end{align*}
and with the primal-dual relationship $R=\Diag(\exp(\gamma f_2))\xi^{(2)}\Diag(\exp(\gamma h_2))$ such that  $-f_2\in\partial F_{\tau_2,b}(R\mathbf{1}_r)$,$-h_2\in\partial G_{\lambda_2}(R^\top\mathbf{1}_m)$.
Finally the dual can be written as 
\begin{align*}
      &\sup_{\lambda_1,\lambda_2} 
      \sup_{f_1,h_1} -F_{\tau_1,a}^{*}(-f_1) - G_{\lambda_1}^{*}(-h_1) - \frac{1}{\gamma}\langle \exp(\gamma(f_1+h_1)),\xi^{(1)}\rangle\\
      &+  \sup_{f_2,h_2} -F_{\tau_2,b}^{*}(-f_2) - G_{\lambda_2}^{*}(-h_2) - \frac{1}{\gamma}\langle \exp(\gamma(f_2+h_2)),\xi^{(2)}\rangle\\
      &-\frac{1}{\gamma}\langle \exp(-\gamma(\lambda_1+\lambda_2)), \xi^{(3)}\rangle + \langle \xi^{(3)},\mathbf{1}\rangle
\end{align*}
and using the definition of $G_{\lambda_1}^{*}(-h_1)$ and $G_{\lambda_2}^{*}(-h_2)$, we obtain the desired dual up to an additive constant $(\langle \xi^{(3)},\mathbf{1}\rangle)$ which does not affect the solution of the problem and conclude the proof.

\subsection{On the Iterations of the Dykstra's Algorithm}
Recall that we propose to consider an alternate maximization scheme to solve ~\eqref{eq-dual-barycenter}. Starting from $h_1^{(0)}=h_2^{(0)}=\mathbf{0}_r$, we apply for $\ell\geq 0$ the following updates (dropping iteration number $k$ in~\eqref{eq-barycenter-ulot} for simplicity):
$$
\begin{aligned}
     f_1^{(\ell+1)}\eqdef\arg\sup_{z} \mathcal{D}(z,h_1^{(\ell)},f_2^{(\ell)},h_2^{(\ell)}), \,f_2^{(\ell+1)}\eqdef\arg\sup_{z} \mathcal{D}(f_1^{(\ell+1)},h_1^{(\ell)},z,h_2^{(\ell)}), \\
     (h_1^{(\ell+1)},h_2^{(\ell+1)})\eqdef \arg\sup_{z_1,z_2} \mathcal{D}(f_1^{(\ell+1)},z_1,f_2^{(\ell+1)},z_2).
\end{aligned}
$$
where 
\begin{align*}
\mathcal{D}(f_1,h_1,f_2,h_2) &=- F_{\tau_1,a}^{*}(-f_1) - \frac{1}{\gamma}\langle e^{\gamma(f_1\oplus h_1)} -1, \xi^{(1)}\rangle 
        - F_{\tau_2,b}^{*}(-f_2) - \frac{1}{\gamma} \langle e^{\gamma(f_2\oplus h_2)} -1, \xi^{(2)}\rangle \\
        &- \frac{1}{\gamma}\langle e^{- \gamma(h_1 + h_2)} - 1,\xi^{(3)}\rangle .
\end{align*}
Let us consider the first update of the scheme that consists in solving
$$  f_1^{(\ell+1)}\eqdef\arg\sup_{z} \mathcal{D}(z,h_1^{(\ell)},f_2^{(\ell)},h_2^{(\ell)})$$

To solve this problem, we again apply the Fenchel-Rockafellar theorem~\citep{rockafellar1970convex} and obtain that
\begin{align*}   
    \sup_{f_1} -F_{\tau_1,a}^{*}(-f_1) - \frac{1}{\gamma}\langle \exp(\gamma(f_1+h_1)),\xi^{(1)}\rangle=\min_s F_{\tau_1,a}(s)+\frac{1}{\gamma} \text{KL}(s|\xi^{(1)}\exp(\gamma h_1))
\end{align*}
and the optimality condition gives that $f_1^{*}$ is solution of the LHS if and only if $s^*$ solves the RHS and belongs to the subdifferential of $f_1\to \exp(\gamma(f_1+h_1)),\xi^{(1)}\rangle$ at $f_1^*$, that is  $s^{*}=\exp(\gamma f_1^*) \odot\xi^{(1)}\exp(\gamma h_1)$. However the RHS problem can can be solved exactly and one obtained that $s^{*}=a^{(\tau_1/(1/\gamma+\tau_1))} \odot  \xi^{(1)}\exp(\gamma h_1)^{(1/(1/1+\gamma\tau_1))}$, therefore when combined with the previous equation on $s^*$ we obtain that
\begin{align*}
    \exp(\gamma f_1^*) =\frac{s^{*}}{\xi^{(1)}\exp(\gamma h_1)}=\left(\frac{a}{\xi^{(1)}\exp(\gamma h_1)}\right)^{\frac{\tau_1}{1/\gamma +\tau_1}},
\end{align*}
Similarly, the solution of 
$  \arg\sup_{z} \mathcal{D}(f_1,h_1,z,h_2)$
is
\begin{align*}
    \exp(\gamma f_2^*) =\left(\frac{b}{\xi^{(2)}\exp(\gamma h_2)}\right)^{\frac{\tau_2}{1/\gamma +\tau_2}} .
\end{align*}
Let us now consider the following optimization problem corresponding to the last update if the alternate maximization scheme, that is
$$ 
(h_1^{(\ell+1)},h_2^{(\ell+1)})\eqdef \arg\sup_{z_1,z_2} \mathcal{D}(f_1^{(\ell+1)},z_1,f_2^{(\ell+1)},z_2).
$$
In fact this problem can be solved directly using simply the first-order condition of optimality that gives the two following equations:
\begin{align*}
    &\exp(\gamma h_1)\odot (\xi^{(1)})^\top \exp(\gamma f_1) - \exp(-\gamma h_1)\odot (\xi^{(3)}) \odot \exp(-\gamma h_2) = 0 \quad \text{and}\\
   & \exp(\gamma h_2)\odot (\xi^{(2)})^\top \exp(\gamma f_2) - \exp(-\gamma h_2)\odot (\xi^{3)}) \odot \exp(-\gamma h_1) = 0
\end{align*}
leading to
\begin{align*}
    g = (\xi^{3)}\odot (\xi^{(1)})^\top \exp(\gamma f_1) \odot(\xi^{(2)})^\top \exp(\gamma f_2) )^{1/3}
\end{align*}
and 
\begin{align*}
  \exp(\gamma h_1) = \frac{g}{(\xi^{(1)})^\top \exp(\gamma f_1)},\quad   \exp(\gamma h_2) = \frac{g}{(\xi^{(2)})^\top \exp(\gamma f_2)}.
\end{align*}

\subsection{Proof of Proposition~\ref{prop-TI}}
Let us consider the following optimization problem 
\begin{align*}
\mathcal{D}_{\textrm{TI}}(\tilde f_1, \tilde h_1, \tilde f_2, \tilde h_2)\eqdef \sup_{\lambda_1,\lambda_2\in\mathbb{R}} \mathcal{D}(\tilde f_1+\lambda_1,\tilde h_1-\lambda_1,\tilde f_2+\lambda_2,\tilde h_2-\lambda_2)
\end{align*}
Therefore we have
\begin{align*}
    &\sup_{\lambda_1,\lambda_2\in\mathbb{R}} \mathcal{D}(\tilde f_1+\lambda_1,\tilde h_1-\lambda_1,\tilde f_2+\lambda_2,\tilde h_2-\lambda_2) \\
   & = -  F_{\tau_1,a}^{*}(-(\tilde f_1 +\lambda_1)) -  F_{\tau_2,b}^{*}(-(\tilde f_2 +\lambda_2))  
    - \frac{1}{\gamma}\langle e^{-\gamma(\tilde h_1+ \tilde h_2)}\odot e^{\gamma(\lambda_1+\lambda_2)}, \xi^{(3)}\rangle + C
\end{align*}
where $C$ does not depends on $\lambda_1$ and $\lambda_2$. Now observe that
\begin{align*}
    F_{\tau_1,a}^{*}(s) &= \sup_{x} \langle x,s\rangle -  \tau_1\text{KL}(s|a)\\
\end{align*}
and by applying the first-order optimality condition, we obtain that $  x^{*} = \exp(s/\tau_1) \odot a$ solves the above optimization problem and 
\begin{align*}
    F_{\tau_1,a}^{*}(s)= \tau_1 \langle \exp(s/\tau_1),a\rangle.
\end{align*}
Similarly, 
\begin{align*}
    F_{\tau_2,b}^{*}(s)= \tau_2 \langle \exp(s/\tau_2),b\rangle,
\end{align*}
Then by appyling the first-order optimality condition we obtain the two following equations
\begin{align*}
    &\exp(-\lambda_1/\tau_1) \langle \exp(-\tilde f_1/\tau_1),a\rangle - \exp(\gamma \lambda_1) \langle \exp(\gamma \lambda_2), \xi^{(3)}\odot \exp(- \gamma(\tilde h_1 + \tilde h_2)) \rangle =0 
    \quad \text{and}\\
    & \exp(-\lambda_2/\tau_2) \langle \exp(-\tilde f_2/\tau_2),b\rangle - \exp(\gamma \lambda_2) \langle \exp(\gamma \lambda_1), \xi^{(3)}\odot \exp(- \gamma(\tilde h_1 + \tilde h_2)) \rangle =0.
\end{align*}
which is equivalent to
\begin{align*}
    \exp\left(\lambda_1 \frac{1/\gamma+\tau_1}{\tau_1/\gamma}\right) &=\frac{ \langle \exp(-\tilde f_1/\tau_1),a\rangle }{\langle \xi^{(3)},  \exp(- \gamma(\tilde h_1 + \tilde h_2)) \rangle }  \exp(-\gamma \lambda_2)\quad \text{and}\\
    \exp\left(\lambda_2 \frac{1/\gamma+\tau_2}{\tau_2/\gamma}\right) &=\frac{ \langle \exp(-\tilde f_2/\tau_2),b\rangle }{\langle \xi^{(3)},  \exp(- \gamma(\tilde h_1 + \tilde h_2)) \rangle }  \exp(-\gamma \lambda_1)
\end{align*}
Then applying $\log$ to the system, we obtain that
\begin{align*}
    \lambda_1 \gamma \frac{1/\gamma + \tau_1}{\tau_1} &= c_1 -\gamma \lambda_2 \quad \text{and}\\
    \lambda_2 \gamma \frac{1/\gamma + \tau_2}{\tau_2} &= c_2 -\gamma \lambda_1
\end{align*}
where 
\begin{align*}
    c_1\eqdef \log\left( \frac{\langle\exp(-  \tilde f_1/\tau_1),a\rangle}{\langle \exp(- \gamma ( \tilde h_1+ \tilde h_2)),\xi^{(3)}\rangle}\right),\quad\text{and}\quad
       c_2\eqdef \log\left( \frac{\langle\exp(-  \tilde f_2/\tau_2),a\rangle}{\langle \exp(- \gamma ( \tilde h_1+ \tilde h_2)),\xi^{(3)}\rangle}\right).
\end{align*}
Finally we obtain a simple linear system and the solution follows.

\subsection{Double Regularizations: Low-rank Structure and Entropy}
Our proposed procedure can be easily extended to the case where one wants to add entropy in addition to the low-rank constraint to solve unbalanced low-rank and entropic optimal transport problems. More precisely, let us consider the general case where one aims at solving  for any $\varepsilon>0$
\begin{equation}
\begin{aligned}
  \text{ULOT}_{r,\varepsilon}(\mu,\nu) \eqdef \min_{(Q,R,g)\in \Pi_r} \underbrace{\langle C,Q \Diag(1/g)R^T\rangle}_{\mathcal{L}_C(Q,R,g)}
   +\underbrace{\tau_1 \text{KL}(Q\mathbf{1}_r|a) + \tau_2 \text{KL}(R\mathbf{1}_r|b) - \varepsilon H(Q,R,g)}_{\mathcal{G}_{a,b,\varepsilon}(Q,R,g)}\; 
\end{aligned}
\end{equation}
where $H(Q,R,g)=H(Q)+H(R)+H(g)$ and $H(p)\eqdef -\sum_i p_i(\log(p_i)-1)$. Note that here, compared to~\eqref{eq-ulot-reformulated}, we have simply add en entropic term to the objective to smooth the matrices $Q,R$ and the barycenter $g$. To solve this problem, we propose to consider the exact same strategy as the one proposed to solve~\eqref{eq-ulot-reformulated} where we slightly modify $\mathcal{G}_{a,b,\varepsilon}$ and explicitely show the dependency w.r.t. $\varepsilon$. Now by applying the linearzation step of $\mathcal{L}_C(Q,R,g)$, we now aim to solve at iteration $k$ the following optimization problem:
\begin{equation}
\label{eq-barycenter-uelgw} 
\begin{aligned}
 (Q_{k+1},R_{k+1},g_{k+1})   \eqdef   \argmin_{\bm{\zeta} \in\Pi_r} \frac{1}{\gamma_k}\text{KL}(\bm{\zeta},\bm{\xi}_k)+ \varepsilon H(\bm{\zeta})+
 \tau_1 \text{KL}(Q\mathbf{1}_r|a) + \tau_2 \text{KL}(R\mathbf{1}_r|b)
 \end{aligned}
\end{equation}
In fact, this problem can be reformulated as a problem of the form~\eqref{eq-barycenter-ulgw} where we simply have to modify $\bm{\xi}_k$ and $\gamma$. Indeed observe that we have
\begin{align*}
    \frac{1}{\gamma} \text{KL}(Q|\xi^{(1)})-\varepsilon H(Q) &=\frac{1}{\gamma_\varepsilon} \text{KL}(Q|\xi_\varepsilon^{(1)})
\end{align*}
where $\gamma_\varepsilon=\frac{1}{1/\gamma + \varepsilon}$ and $\xi_\varepsilon^{(1)}:=(\xi^{(1)})^{\gamma_\varepsilon/\gamma}$. Therefore we obtain that
\begin{align*}
 &\argmin_{\bm{\zeta} \in\Pi_r} \frac{1}{\gamma}\text{KL}(\bm{\zeta},\bm{\xi})+ \varepsilon H(\bm{\zeta})+
 \tau_1 \text{KL}(Q\mathbf{1}_r|a) + \tau_2 \text{KL}(R\mathbf{1}_r|b)\\
 &= \argmin_{\bm{\zeta} \in\Pi_r} \frac{1}{\gamma_\varepsilon}\text{KL}(\bm{\zeta},\bm{\xi}_\varepsilon)+
 \tau_1 \text{KL}(Q\mathbf{1}_r|a) + \tau_2 \text{KL}(R\mathbf{1}_r|b)
\end{align*}
where $\bm{\xi}_\varepsilon:=(\xi_\varepsilon^{(1)},\xi_\varepsilon^{(2)},\xi_\varepsilon^{(3)})$. Therefore the entropic version of our problem can be solved using the exact same solver as the one proposed in the main paper where only simple updates of the gradient-step $\gamma$ and the kernels $\bm{\xi}$ are required at each iteration. We summarize the proposed algorithm below.

\begin{algorithm}[H]
\SetAlgoLined
\textbf{Inputs:} $C, a, b, \varepsilon, \gamma_0, \tau_1,\tau_2,\delta$\\
$Q,R,g \gets \text{Initialization as proposed in~\citep{scetbon2022lot}}$\\
\Repeat{$\Delta((Q,R,g),(\tilde Q,\tilde R,\tilde g),\gamma)<\delta$}{
    $\tilde{Q}=Q,~~\tilde{R}=R,~~\tilde{g}=g,\\
    \nabla_Q = CR \Diag(1/g),~~\nabla_R = C^\top Q \Diag(1/g),\\
    \omega \gets \mathcal{D}(Q^TCR),~~\nabla_g = - \omega / g^2,\\
    \gamma \gets \gamma_0 / \max(\|\nabla_Q\|_{\infty}^2,\|\nabla_R\|_{\infty}^2, \|\nabla_g\|_{\infty}^2),\\
    \gamma \gets \frac{1}{1/\gamma + \varepsilon}\\
    \xi^{(1)}\gets Q \odot \exp(-\gamma\nabla_Q),~\xi^{(2)}\gets R\odot \exp(-\gamma\nabla_R),~\xi^{(3)}\gets g\odot \exp(-\gamma\nabla_g),\\
    \xi^{(1)}\gets (\xi^{(1)})^{\gamma_\varepsilon/\gamma},~\xi^{(2)}\gets (\xi^{(2)})^{\gamma_\varepsilon/\gamma},~\xi^{(3)}\gets (\xi^{(3)})^{\gamma_\varepsilon/\gamma},\\
    Q,R,g\gets \text{ULR-Dykstra}(a,b,\bm{\xi},\gamma,\tau_1,\tau_2,\delta)~(\text{Alg.~\ref{alg-dykstra-ulot})}$
  }
\textbf{Result:} $Q,R,g$
\caption{$\text{ULOT}_\varepsilon(C,a,b,r,\gamma_0,\tau_1,\tau_2,\delta)$ \label{alg-prox-uelot}}
\end{algorithm}

\end{document}